\pdfoutput=1

\PassOptionsToPackage{table}{xcolor}
\documentclass[10pt,twocolumn,letterpaper]{article}

\usepackage{adjustbox}
\usepackage[pagenumbers]{cvpr} %

\usepackage{listings}
\lstdefinestyle{chdprompt}{
  basicstyle=\ttfamily\fontsize{8}{9}\selectfont,
  columns=fullflexible,
  keepspaces=true,
  breaklines=true,
  breakatwhitespace=true,
  showstringspaces=false,
  backgroundcolor=\color{black!5},
  rulecolor=\color{black!25},
  frame=single,
  framerule=0.4pt,
  framesep=10pt,
  xleftmargin=10.4pt,
  xrightmargin=10.4pt,
  aboveskip=0pt,
  belowskip=0pt,
  captionpos=t,
  abovecaptionskip=6pt,
  belowcaptionskip=8pt
}
\lstdefinestyle{chdpromptcompact}{
  style=chdprompt,
  framesep=6pt,
  xleftmargin=6.4pt,
  xrightmargin=6.4pt,
  escapeinside={(*@}{@*)}
}
\usepackage{xspace}
\newcommand{\gptsol}{GPT-6.1 Sol\xspace}
\newcommand{\opus}{Claude Opus~5.5\xspace}
\newcommand{\gemini}{Gemini~4 Argon\xspace}
\newcommand{\medicalnet}{MedicalNet 3D ResNet-18\xspace}
\newcommand{\ci}[2]{#1{\tiny\,$\pm$\,#2}} %

\usepackage{booktabs} 
\usepackage{multirow} 
\usepackage[table]{xcolor}
\definecolor{rowalt}{gray}{0.86}  %
\definecolor{rowbest}{rgb}{0.78,0.84,0.95} %
\usepackage{iftex}
\ifXeTeX
  \usepackage{fontspec}
\fi

\definecolor{cvprblue}{rgb}{0.21,0.49,0.74}
\usepackage[pagebackref,breaklinks,colorlinks,allcolors=cvprblue]{hyperref}

\usepackage{fontawesome5}
\definecolor{repoblue}{rgb}{0.07,0.24,0.47}
\newcommand{\repourl}{%
  \href{https://github.com/Maitri-Labs-Amrita/CHD-Bench}{\textcolor{repoblue}{\faGithub\ \texttt{github.com/Maitri-Labs-Amrita/CHD-Bench}}}}

\def\confName{CVPR}
\def\confYear{2027}

\title{CHD-Bench: Diagnosing Congenital Heart Disease through Interactive Examination}

\author{
Gautham Krishnan P$^{1}$ \quad Shrisharanyan Vasu$^{1}$ \quad Aniketh Vijesh$^{1}$ \quad Rajagopalan Rameshbabu$^{1}$\\
Harikrishnan Anil Maya$^{2}$ \quad Sarin Xavier$^{2}$ \quad Mahesh Kappanayil$^{2}$ \quad Gilad Gressel$^{1,*}$\\[0.6em]
{\normalsize $^{1}$School of Engineering, Amrita Vishwa Vidyapeetham, Amritapuri, India}\\
{\normalsize $^{2}$Department of Pediatric Cardiology, Amrita Institute of Medical Sciences, Kochi, India}\\[0.3em]
{\small $^{*}$Corresponding author: \texttt{gilad.gressel@am.amrita.edu}}\\[0.7em]
{\large \repourl}
}

\begin{document}

\twocolumn[{%
    \renewcommand\twocolumn[1][]{#1}%
    \maketitle

    \centering
    \includegraphics[width=\textwidth]{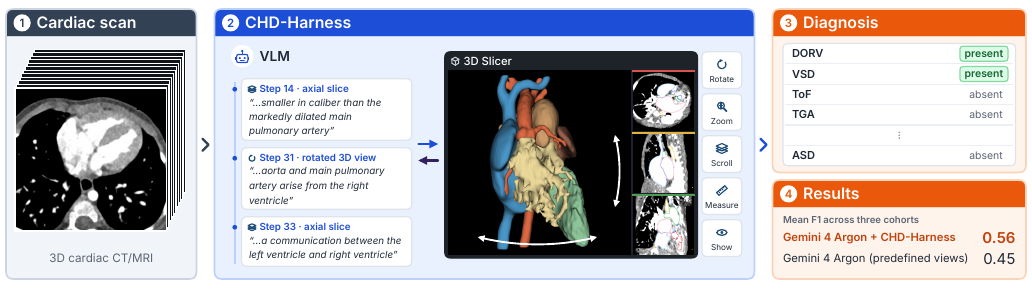}
    \par

    \captionsetup{hypcap=false}
    \captionof{figure}{
        \textbf{Interactive CHD diagnosis with CHD-Harness.}
        A VLM examines a 3D cardiac scan in 3D Slicer step by step, choosing its own views while describing what it sees, and reports every CHD type as present or absent.
    }
    \label{fig:teaser}

    \vspace{1em}
}]

\begin{abstract}
Diagnosis of congenital heart disease (CHD) requires understanding the spatial relationships between cardiac structures, which clinicians assess by scrolling through CT or cardiac magnetic resonance imaging (MRI) slices and rotating 3D reconstructions. Although automated methods have advanced cardiac segmentation, identifying congenital heart defects from volumetric scans remains an open challenge. The two existing approaches rely on handcrafted diagnostic rules limited to static anatomical abnormalities. We introduce CHD-Bench, the first benchmark for interactive CHD diagnosis, defining a multi-label task and evaluation protocol over 247 CT and MRI scans from three cohorts covering 38 CHD types. We also present CHD-Harness, open-source software that runs a VLM in a 3D Slicer environment through 22 tools, manages each examination, and records the resulting diagnosis and supporting views. We evaluate off-the-shelf medical, open weight, and closed weight VLMs and compare them with a 3D CNN classifier and published rule-based results. \gemini with CHD-Harness achieves the highest macro F1 across all three cohorts (0.63, 0.63, and 0.43). Allowing for interactive examination raises its macro F1 by up to 20 points over static views. These results establish CHD-Bench and CHD-Harness as a foundation for developing and evaluating VLMs for interactive CHD diagnosis.
\end{abstract}

\section{Introduction}
\label{sec:intro}

\begin{figure*}[t]
    \centering
    \includegraphics[width=\textwidth]{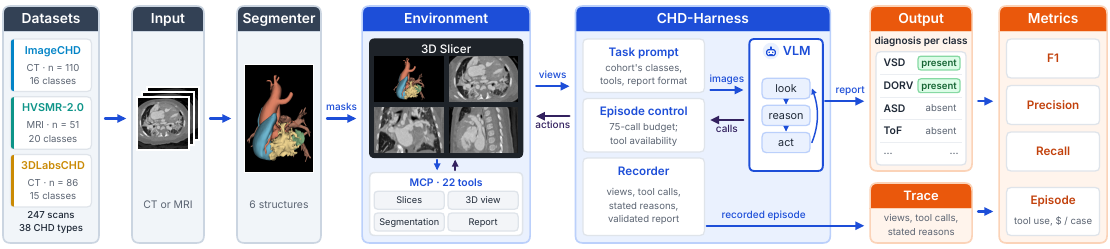}
    \caption{\textbf{Overview of CHD-Bench and CHD-Harness.}
    CHD-Bench defines the diagnostic task and evaluation protocol. CHD-Harness connects a VLM to a 3D Slicer environment through 22 tools, allowing it to interactively examine CT and MRI scans and produce multi-label diagnoses.}

    \label{fig:overview}
\end{figure*}

Congenital heart disease (CHD) is the most common major structural birth defect, affecting approximately 9 in every 1{,}000 live births~\cite{vanderlinde2011}. Its treatment depends on which defects are present and how the cardiac chambers and great vessels connect~\cite{bierhals2014}. Associated vascular anomalies can alter the surgical approach even when the principal diagnosis is established~\cite{goo2018coronary}. While simpler defects can often be assessed using two-dimensional echocardiography, complex CHD requires computed tomography (CT) or cardiac magnetic resonance imaging (MRI) to understand the spatial relationships between cardiac structures~\cite{baumgartner2021esc,secinaro2022sicp,goo2020advanced}. Clinicians examine these volumetric scans by scrolling through 2D slices and manipulating multiplanar and 3D views to identify abnormalities and trace anatomical connections. This process requires considerable expertise and spatial reasoning, particularly in cases with severely altered cardiac anatomy. Although producing 3D reconstructions previously required substantial manual segmentation~\cite{imagechd}, this process is now largely automated~\cite{nnunet,stacom}. However, both human and automated diagnosis of complex CHD from these scans remains challenging.

To our knowledge, only two published studies classify CHD types from volumetric scans, and both combined segmentation with handcrafted rules for anatomical connections~\cite{imagechd,XU2023102953}. Such rules encode only patterns defined in advance, whereas CHD presents in many forms and combinations~\cite{XU2023102953}. The large combinatorial space makes it challenging to train models to classify CHD directly from the volumes.

Clinicians diagnose CHD through iterative examination of cardiac structures and their connections~\cite{anderson1984sequential,lapierre2010segmental}. They inspect the anatomy from different viewpoints, integrate observations across views, and follow up on suspected abnormalities until they reach a diagnosis. This process requires multiple steps of anatomical reasoning, with findings guiding subsequent inspection. Recent advances in multi-step reasoning and tool use in large language models~\cite{gemini2025gemini25} motivate a different approach to automation: giving a general-purpose model direct access to an imaging environment and allowing it to choose how to examine the scan. Can a vision-language model (VLM) diagnose CHD by examining a cardiac scan as a clinician does?

To answer this question, we introduce \textbf{CHD-Bench}, a benchmark that defines the scans and diagnostic labels, the multi-label task, and a standardized evaluation protocol. It comprises 247 CT and MRI scans from three cohorts, spanning 38 CHD types. We also introduce \textbf{CHD-Harness}, model-independent software for running VLMs on this benchmark (Fig.~\ref{fig:teaser}). It connects a VLM to a 3D Slicer environment~\cite{fedorov2012slicer,slicer_website} through a Model Context Protocol (MCP) server and manages the examination loop. The VLM chooses among 22 tools to navigate slices, manipulate 3D views, inspect segmentations, and measure anatomy. CHD-Harness records the tool calls and supporting image views, making each examination traceable (Fig.~\ref{fig:overview}).

We evaluate 12 models, including open and closed weight VLMs, specialized medical VLMs, and a 3D CNN classifier, alongside ablation studies that assess the contributions of interactive examination and segmentation. The VLMs are evaluated off the shelf, without CHD-specific fine-tuning, labeled demonstrations, or handcrafted diagnostic rules.

\textbf{An off-the-shelf frontier VLM can identify complex heart defects by choosing its own views in an existing medical imaging environment.} \gemini with CHD-Harness achieves the highest macro F1 across all three cohorts, reaching 0.63 on both public cohorts and 0.43 on the private cohort. Its mean macro F1 is 0.56, compared with 0.31 for the strongest open weight VLM and 0.26 for the MedicalNet classifier.
Interactive examination improves macro F1 by 4 to 20 points over static views, while removing segmentation or the 3D view reduces mean macro F1 by only one point. On ImageCHD, \gemini achieves a macro recall of 0.72, matching the published rule-based method~\cite{imagechd}, but remaining below the 0.91 achieved by clinical readers on the same dataset. Importantly, this performance is achieved without CHD-specific training or diagnostic rules, suggesting that advances in general-purpose VLMs will further narrow the gap to clinical performance.

\begin{figure*}[t]
\centering
 \includegraphics[width=\linewidth]{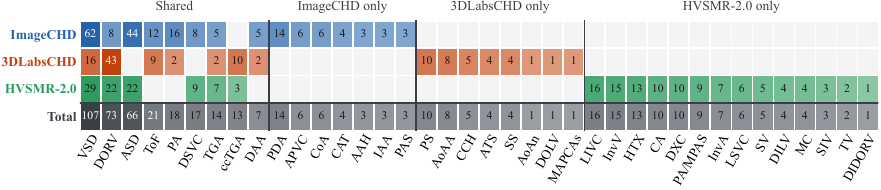}
\caption{\textbf{Distribution of the 38 CHD types in CHD-Bench.} Each cell gives the number of positive cases of a class in a cohort, and the last row gives the total. Empty cells mark classes not labeled in a cohort. The class abbreviations are defined in Table~\ref{tab:abbrev} (Appendix~\ref{sec:suppl_data}).}
\label{fig:class_dist}
\end{figure*}

CHD is a demanding test case for a broader approach: connecting general-purpose VLMs to the medical imaging software that clinicians already use. CHD-Bench and CHD-Harness provide a common setting for evaluating successive VLMs as they become available. We release both to support progress toward useful diagnostic assistance, with CHD-Harness designed to extend to other medical imaging tasks by changing only the input scans and task prompt. Our contributions are threefold:
\begin{itemize}
  \item \textbf{CHD-Bench.} The first benchmark for interactive CHD diagnosis, defining the task and evaluation protocol over 247 CT and MRI scans across three cohorts spanning 38 CHD types.
  \item \textbf{CHD-Harness.} Open-source software that connects a tool-calling VLM to a 3D Slicer environment through a custom MCP server, manages the examination loop, and records every examination.
  \item \textbf{Evaluation.} Evidence that off-the-shelf frontier VLMs can identify CHD through interactive examination, supported by comparisons across 12 models and ablations of the viewing tools and segmentation.

\end{itemize}
\section{Related Work}
\label{sec:related_works}

\noindent\textbf{Automated CHD diagnosis.}
Learning-based CHD diagnosis has primarily focused on 2D modalities, including fetal ultrasound~\cite{arnaout2021ensemble}, echocardiography~\cite{cheng2024echo}, and chest radiographs~\cite{toba2020cxr}. However, diagnosing complex CHD requires understanding the spatial relationships between cardiac chambers and great vessels, which are best assessed through volumetric imaging~\cite{goo2020advanced,bierhals2014}. Most prior work on cardiac CT and MRI has therefore focused on anatomical segmentation~\cite{byrne2016segmentation,pace,xu2019accurate,hvsmr2,kong2024sdf4chd}, with only two studies addressing diagnosis directly. Xu et al.~\cite{imagechd,XU2023102953} classify CHD using segmentation-derived anatomical features and handcrafted diagnostic rules. However, their diagnostic implementations are not publicly available, preventing independent reproduction. In contrast, \textbf{CHD-Bench establishes a standardized, reproducible evaluation across 247 scans, three cohorts, two imaging modalities, and 38 CHD types}. Together with CHD-Harness, it enables systematic evaluation of general-purpose VLMs that diagnose CHD through interactive anatomical examination, without task-specific fine-tuning or handcrafted diagnostic rules.

\noindent\textbf{Medical vision-language models.} Foundational~\cite{gemini2025gemini25} and medical VLMs~\cite{medgemma,vilam3,medmo,octomed} answer questions about medical images and generate reports, and these capabilities are measured across organs and imaging modalities by multiple benchmarks~\cite{medcmr,omnibrainbench,gastricx}. Most of these studies operate on images selected in advance, often single 2D slices, so the model never decides which part of a volume to examine. Volumetric models~\cite{m3d,radfm} encode the whole scan once, at a fixed resolution and token budget, so every structure is represented at the same level of detail regardless of the question, and they output a single text answer without any accompanying visual evidence. In cardiac imaging, Li et al.~\cite{cmrrd} introduced CMR-RD, a VLM adapted to address long-tailed disease distributions in cardiac MRI. CHD-Bench complements this work by evaluating models that select their own views of the full volume.

\noindent\textbf{Tool-using agents.}
Recent VLMs extend visual reasoning through actions such as cropping, high-resolution inspection, and code execution~\cite{codevision,lu2026agenticrl,sensesearch}, while agents have also been developed to answer 3D spatial queries through programmatic reasoning~\cite{vadar}. Agentic benchmarks increasingly involve interaction with professional software, including Blender~\cite{blendergym}, ComfyUI~\cite{comfybench}, and desktop applications~\cite{prosoftarena}, while MCP has enabled model interaction with scientific visualization tools such as ParaView~\cite{paraviewmcp}. In medicine, agents have been used to coordinate specialist models~\cite{moa,ustri,radagent}, refine segmentations~\cite{ibisagent}, examine dental radiographs~\cite{oralgptplus}, operate clinical software~\cite{carepilot}, and optimize imaging workflows~\cite{simpleagents}. Closest to our work, CT-Flow~\cite{ctflow} and ACTOR~\cite{actor} enable agents to examine CT scans through viewing tools, while MedOpenClaw~\cite{medopenclaw} introduces an agent runtime for navigating medical imaging studies through 2D slice-based interactions. 
However, these approaches do not address interactive exploration of 3D cardiac anatomy, where diagnosis depends on understanding the complex spatial relationships between chambers and great vessels rather than detecting localized findings. To our knowledge, \textbf{CHD-Harness is the first system to enable VLMs to interactively manipulate and examine 3D cardiac anatomy for CHD diagnosis}, allowing them to independently explore anatomical structures, determine their own examination strategy, and gather supporting evidence for their diagnoses.

\section{CHD-Bench}
\label{sec:chd_bench}
Given a 3D cardiac scan (CT or MRI), a model must diagnose congenital heart disease. Each cohort defines a list of CHD classes $\mathcal{C}$, and for every class $c \in \mathcal{C}$ the model predicts whether that defect is present, giving a binary vector $\hat{\mathbf{y}} \in \{0,1\}^{|\mathcal{C}|}$. A case with no predicted defect is marked as normal. The task is multi-label, since a case can have several defects at once. In the interactive setting, models examine the scan through the tool interface described in Sec.~\ref{sec:chd_harness} and submit a diagnostic report within a tool-call budget. Non-interactive baselines are evaluated on the same diagnostic labels and metrics.

\subsection{Dataset}
CHD-Bench combines three cohorts spanning two imaging modalities. ImageCHD~\cite{imagechd} is a public dataset of 110 cardiac CT scans with 16 classes. HVSMR-2.0~\cite{hvsmr2} is a public dataset of 60 MRI scans with 20 classes, of which 51 scans have eligible CHD labels. 3DLabsCHD is a private clinical dataset of 86 cardiac CT scans with 15 classes, collected at Amrita Institute of Medical Sciences, Kochi, India. In total, we evaluate 247 scans covering 38 distinct CHD types, including rare and complex abnormalities such as criss-cross heart, scimitar syndrome, and double-outlet left ventricle. The label distribution is highly long-tailed, as demonstrated in Fig.~\ref{fig:class_dist}, where there are  several defects represented by only 1 to 3 cases. Many patients have multiple co-occurring defects, with up to 9 diagnoses in a single scan, making this a challenging multi-label classification task. Further details on dataset acquisition, clinical annotation, label harmonization, and exclusion criteria are provided in Appendix~\ref{sec:suppl_data}.

\subsection{Evaluation Metrics}
We evaluate multi-label CHD classification using precision, recall, and F1 score, computed independently for each diagnostic class and macro-averaged within each cohort. Precision measures the proportion of predicted defects that are correct, while recall measures the proportion of actual defects identified. F1 summarizes the balance between the two. We use macro-averaging to give equal weight to each CHD type, regardless of its prevalence, reflecting the importance of detecting rare as well as common congenital abnormalities. Because the cohorts differ in their diagnostic classes and distributions, we evaluate them separately. We report 95\% confidence intervals estimated from 2{,}000 patient-level bootstrap resamples.
\section{CHD-Harness}
\label{sec:chd_harness}
CHD-Harness is the software used to run interactive VLM examinations on CHD-Bench. It combines an episode runner with a 3D Slicer environment and its MCP server (Fig.~\ref{fig:overview}). The VLM receives rendered views of the scan, chooses what to inspect, and requests tools to change the view. The runner maintains the interaction history, executes the requested tools through the server, and records the examination until the model submits a valid report or reaches the tool-call limit.

\subsection{3D Slicer Environment}
The viewing environment uses 3D Slicer~\cite{fedorov2012slicer} (v5.10) to load, display, and manipulate each scan. A custom MCP server exposes these operations as tools that the VLM can call. The scan is loaded as an anonymized volume, so the model never sees the case identifier or any clinical information. After each tool call, the model receives the rendered images of the views that the call changed (3D, axial, sagittal, or coronal, each $512 \times 512$ pixels) together with a JSON description of the viewer state. We include a segmentation model in the pipeline to simulate the workflow of a clinician who has access to a 3D model of the heart. This model provides a segmentation of six cardiac structures (the left and right atria, the left and right ventricles, the aorta, and the pulmonary artery), displayed as hollow, color-coded surfaces. The predicted masks can contain errors, so the VLM is told to treat them as a guide rather than ground truth. Every episode starts from the same state, in which the 3D view shows the segmented heart from the front without the chest.

\subsection{MCP Server and Tools}
\begin{figure}[t]
\centering
\includegraphics[width=\columnwidth]{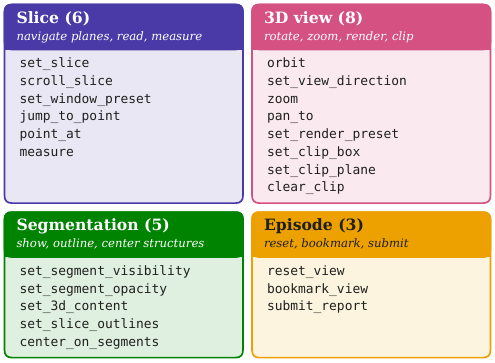}
\caption{\textbf{The 22 tools exposed by the 3D Slicer MCP server.}}
\label{fig:tools}
\end{figure}

Figure~\ref{fig:tools} lists the 22 tools exposed by the MCP server in four groups, and every position is given in patient coordinates in millimeters. \textit{Slice tools} move the axial, sagittal, and coronal views, set the grayscale window, and read the intensity or measure a distance at a chosen pixel, so the model can follow a vessel or a septum across planes. \textit{3D view tools} move the camera, change the render preset, and clip away tissue, so the model can look inside the heart. \textit{Segmentation tools} show, hide, or fade each segmented structure and draw its outline on the slices, so the model can trace how the chambers and great arteries connect. \textit{Episode tools} reset the view, bookmark the current view as evidence, and submit the final report.

\subsection{Examination Loop}
\noindent\textbf{System prompt.}
The model is instructed to act as an expert in cardiac imaging, given a list of possible defects (per cohort) and told to determine which defects are present using only the scan. We restrict predictions to each cohort's annotated classes because CHD labels are not necessarily exhaustive. A defect absent from the annotations may still be present in the scan, making predictions outside the labeled set impossible to evaluate reliably. Thus, the class list defines the scope of evaluation rather than providing diagnostic guidance. The prompt also describes the available tools and coordinate conventions but provides no diagnostic criteria, examples, or predefined examination strategy. The model independently determines which anatomical structures to inspect and how to reach a diagnosis. The full prompt is given in Listing~\ref{lst:system-prompt} (Appendix~\ref{sec:suppl_prompt}).

\noindent\textbf{Inputs and outputs.} The first message contains the initial observation, with all four views and the geometry of the volume. At every later step, the model receives the result of its last tool call, which consists of a JSON block with the viewer state and the number of tool calls left, followed by the rendered images of the changed views. In return, the model describes what it saw and what it will check next, and then issues one or more tool calls. The model is stateless between turns, so every request contains the full interaction history, including all earlier images.

\noindent\textbf{Episode control.}
The episode runner enforces a budget of 75 tool calls, and each examination concludes with a structured multi-label diagnostic report. The MCP server validates the report for consistency, including enforcing that a \textit{Normal} classification is mutually exclusive with any CHD diagnosis. Throughout the examination, the model can bookmark relevant image views as supporting evidence for individual diagnoses. CHD-Harness records the complete interaction history, including the model's reasoning, tool calls, and rendered images. The saved views and interaction history support subsequent clinician review of the diagnosis.

\begin{table*}[t]
\centering
\caption{\textbf{Performance on CHD-Bench.}
Interactive VLMs use CHD-Harness with all tools and predicted segmentations, and baselines are medical VLMs and a CNN classifier that do not use tools.
We report macro F1, precision (P), and recall (R) for each cohort with 95\% confidence intervals from 2000 bootstrap resamples.
Tool calls are the mean per case, baselines run locally and have no API cost, and the best result in each column is in bold.}
\label{tab:main}
\begingroup
\footnotesize
\setlength{\tabcolsep}{2.2pt}
\begin{adjustbox}{max width=\textwidth}
\begin{tabular}{@{}lcccccccccccc@{}}
\toprule
 &
 & \multicolumn{3}{c}{ImageCHD (CT, $n{=}110$)}
 & \multicolumn{3}{c}{HVSMR-2.0 (MRI, $n{=}51$)}
 & \multicolumn{3}{c}{3DLabsCHD (CT, $n{=}86$)}
 & & \\
\cmidrule(lr){3-5}
\cmidrule(lr){6-8}
\cmidrule(lr){9-11}
Model & Mean F1\,$\uparrow$
      & F1\,$\uparrow$ & P\,$\uparrow$ & R\,$\uparrow$
      & F1\,$\uparrow$ & P\,$\uparrow$ & R\,$\uparrow$
      & F1\,$\uparrow$ & P\,$\uparrow$ & R\,$\uparrow$
      & Tool calls & \$/case\,$\downarrow$ \\
\midrule
\multicolumn{13}{@{}l}{\textit{Baselines}} \\
NV-Reason-CT
 & \ci{0.03}{0.01}
 & \ci{0.04}{0.03} & \ci{0.07}{0.05} & \ci{0.08}{0.02}
 & \ci{0.03}{0.02} & \ci{0.03}{0.01} & \ci{0.05}{0.03}
 & \ci{0.02}{0.02} & \ci{0.01}{0.02} & \ci{0.02}{0.03}
 & -- & -- \\
\rowcolor{rowalt}M3D-LaMed
 & \ci{0.04}{0.00}
 & \ci{0.04}{0.01} & \ci{0.03}{0.01} & \ci{0.06}{0.00}
 & \ci{0.03}{0.01} & \ci{0.03}{0.01} & \ci{0.05}{0.00}
 & \ci{0.04}{0.01} & \ci{0.04}{0.01} & \ci{0.05}{0.01}
 & -- & -- \\
ClinFusion-8B
 & \ci{0.07}{0.03}
 & \ci{0.13}{0.05} & \ci{0.17}{0.05} & \ci{0.19}{0.08}
 & \ci{0.09}{0.09} & \ci{0.32}{0.30} & \ci{0.10}{0.07}
 & \ci{0.00}{0.00} & \ci{0.00}{0.00} & \ci{0.00}{0.00}
 & -- & -- \\
\rowcolor{rowalt}MedGemma-27B$^\dagger$
 & \ci{0.11}{0.03}
 & \ci{0.09}{0.03} & \ci{0.10}{0.03} & \ci{0.14}{0.05}
 & \ci{0.14}{0.05} & \ci{0.17}{0.08} & \ci{0.16}{0.06}
 & \ci{0.11}{0.05} & \ci{0.11}{0.06} & \ci{0.14}{0.09}
 & -- & -- \\
MedicalNet
 & \ci{0.26}{0.03}
 & \ci{0.23}{0.05} & \ci{0.20}{0.05} & \ci{0.28}{0.07}
 & \ci{0.32}{0.06} & \ci{0.27}{0.05} & \ci{0.44}{0.07}
 & \ci{0.24}{0.07} & \ci{0.19}{0.05} & \ci{0.36}{0.10}
 & -- & -- \\
\midrule
\multicolumn{13}{@{}l}{\textit{Open weight VLMs}} \\
Qwen3.8-27B
 & \ci{0.18}{0.04}
 & \ci{0.17}{0.07} & \ci{0.27}{0.09} & \ci{0.17}{0.07}
 & \ci{0.19}{0.05} & \ci{0.29}{0.10} & \ci{0.21}{0.05}
 & \ci{0.17}{0.07} & \ci{0.19}{0.08} & \ci{0.30}{0.10}
 & 38.6 & \$0.26 \\
\rowcolor{rowalt}DeepSeek V4.1 Flash
 & \ci{0.19}{0.03}
 & \ci{0.15}{0.05} & \ci{0.20}{0.08} & \ci{0.15}{0.05}
 & \ci{0.24}{0.05} & \ci{0.36}{0.08} & \ci{0.25}{0.06}
 & \ci{0.17}{0.05} & \ci{0.17}{0.06} & \ci{0.29}{0.10}
 & 69.8 & \textbf{\$0.21} \\
Muse Spark Open
 & \ci{0.23}{0.04}
 & \ci{0.16}{0.05} & \ci{0.17}{0.05} & \ci{0.19}{0.06}
 & \ci{0.37}{0.08} & \ci{0.39}{0.08} & \ci{0.39}{0.11}
 & \ci{0.16}{0.05} & \ci{0.14}{0.07} & \ci{0.34}{0.10}
 & 47.3 & \$0.49 \\
\rowcolor{rowalt}Kimi K3.1 Pro
 & \ci{0.31}{0.05}
 & \ci{0.26}{0.06} & \ci{0.36}{0.08} & \ci{0.30}{0.07}
 & \ci{0.39}{0.09} & \ci{0.51}{0.10} & \ci{0.40}{0.09}
 & \ci{0.28}{0.10} & \ci{0.30}{0.11} & \ci{0.45}{0.11}
 & 59.4 & \$1.16 \\
\midrule
\multicolumn{13}{@{}l}{\textit{Closed weight VLMs}} \\
\opus
 & \ci{0.49}{0.05}
 & \ci{0.57}{0.09} & \ci{0.59}{0.11} & \ci{0.65}{0.09}
 & \ci{0.57}{0.08} & \ci{0.59}{0.08} & \ci{0.59}{0.10}
 & \ci{0.32}{0.08} & \ci{0.31}{0.08} & \ci{0.49}{0.12}
 & 70.5 & \$2.91 \\
\rowcolor{rowalt}\gptsol
 & \ci{0.49}{0.05}
 & \ci{0.58}{0.08} & \ci{0.59}{0.09} & \ci{0.60}{0.09}
 & \ci{0.58}{0.08} & \ci{0.60}{0.09} & \ci{0.64}{0.08}
 & \ci{0.32}{0.10} & \ci{0.34}{0.11} & \ci{0.40}{0.13}
 & 58.6 & \$0.74 \\
\rowcolor{rowbest}\gemini
 & \ci{\textbf{0.56}}{0.05}
 & \ci{\textbf{0.63}}{0.08} & \ci{\textbf{0.62}}{0.09} & \ci{\textbf{0.72}}{0.08}
 & \ci{\textbf{0.63}}{0.08} & \ci{\textbf{0.63}}{0.09} & \ci{\textbf{0.69}}{0.10}
 & \ci{\textbf{0.43}}{0.08} & \ci{\textbf{0.38}}{0.08} & \ci{\textbf{0.68}}{0.13}
 & 59.6 & \$1.31 \\
\bottomrule
\end{tabular}
\end{adjustbox}
\par\smallskip
{\raggedright $^\dagger$MedGemma-27B does not produce a valid report for 56 of the 247 cases, which are excluded from its metrics.\par}
\endgroup
\end{table*}

\section{Experiments}
\label{sec:experiments}

\subsection{Experimental Setup}
\label{sec:setup}

\noindent\textbf{Segmentation model.} The 3D Slicer environment displays segmentation masks predicted by a SwinUNETR~\cite{swinunetr}, which we train on the six cardiac structures described in Sec.~\ref{sec:chd_harness}.
Training uses three-fold cross-validation over the scans of all three cohorts, so every mask the VLM sees comes out-of-fold, from a segmentation model that was not trained on that scan. On the held-out folds, the mean Dice over the six structures is 0.77 (0.82 on ImageCHD, 0.73 on HVSMR-2.0, and 0.73 on 3DLabsCHD). The training details are given in Appendix~\ref{sec:suppl_seg}.

\noindent\textbf{Models.} We evaluate three groups of VLMs, none of which is trained on CHD labels. We use each model's highest reasoning setting and ensure that no response is limited by output length. The closed weight models \gptsol~\cite{gptsol}, \gemini~\cite{geminiargon}, and \opus~\cite{claudeopus} and the open weight models Kimi K3.1 Pro~\cite{kimi}, Muse Spark Open~\cite{muse}, DeepSeek V4.1 Flash~\cite{deepseek}, and Qwen3.8-27B~\cite{qwen38} examine scans interactively with CHD-Harness. We access \opus through Anthropic's API and all other interactive models through OpenRouter.
The medical VLMs ClinFusion-8B~\cite{clinfusion}, M3D-LaMed~\cite{m3d}, and NV-Reason-CT~\cite{nvreasonct} take a whole 3D volume as input, so we give each model the full scan in a single pass and ask it for the same multi-label diagnosis. MedGemma-27B~\cite{medgemma} cannot take 3D volumes as input or call tools, so we provide it with a fixed set of 2D slices and 3D renderings from each scan. MedGemma-27B also does not produce a valid report for 56 of the 247 cases, so we exclude these cases from its metric calculation. All four models run locally. Complete technical details, including the full-volume prompt (Listing~\ref{lst:volume-prompt}), are given in Appendix~\ref{sec:suppl_baselines}.

\noindent\textbf{Volumetric classifier.} We also evaluate a volumetric classifier built on MedicalNet~\cite{chen2019med3d}, which predicts the labels from the whole scan without examining it interactively. Many CHD types have only a few examples, so we keep the pretrained \medicalnet frozen rather than fine-tune it, and train one class-balanced linear SVM per label on its pooled features. The SVMs are trained separately for each cohort, with the same three cross-validation folds as the segmentation model. The implementation details of all baselines are given in Appendix~\ref{sec:suppl_baselines}.

\noindent\textbf{Evaluation protocol.} We evaluate each model once per scan and score each cohort separately with the metrics of Sec.~\ref{sec:chd_bench}, together with the mean macro F1 over the three cohorts. The CHD-Bench evaluation code and the CHD-Harness implementation, with the system prompts, tool definitions, and scoring script, are publicly available at \url{https://github.com/Maitri-Labs-Amrita/CHD-Bench} (Appendix~\ref{sec:suppl_code}). We obtain all results with CHD-Bench v0.1.0 and slicer-eval-mcp v0.1.0, the open-source MCP server through which the model controls 3D Slicer (\url{https://github.com/Maitri-Labs-Amrita/slicer-eval-mcp}).

\begin{table*}[t]
\centering
\caption{\textbf{Ablations of \gemini with CHD-Harness.}
\textit{Static views} removes all tools and gives the model a fixed set of slices and 3D views.
\textit{No 3D} removes the 3D view and the tools that act on it.
\textit{No segmentation} removes the predicted segmentation masks and the five segmentation tools.
\textit{Full configuration} uses all tools and predicted segmentations.
Metrics and notation follow Table~\ref{tab:main}.}
\label{tab:ablation}
\begingroup
\footnotesize
\setlength{\tabcolsep}{2.2pt}
\begin{adjustbox}{max width=\textwidth}
\begin{tabular}{@{}lcccccccccccc@{}}
\toprule
 &
 & \multicolumn{3}{c}{ImageCHD (CT, $n{=}110$)}
 & \multicolumn{3}{c}{HVSMR-2.0 (MRI, $n{=}51$)}
 & \multicolumn{3}{c}{3DLabsCHD (CT, $n{=}86$)}
 & & \\
\cmidrule(lr){3-5}
\cmidrule(lr){6-8}
\cmidrule(lr){9-11}
Setting & Mean F1\,$\uparrow$
      & F1\,$\uparrow$ & P\,$\uparrow$ & R\,$\uparrow$
      & F1\,$\uparrow$ & P\,$\uparrow$ & R\,$\uparrow$
      & F1\,$\uparrow$ & P\,$\uparrow$ & R\,$\uparrow$
      & Tool calls & \$/case\,$\downarrow$ \\
\midrule
Static views
 & \ci{0.45}{0.05}
 & \ci{0.43}{0.07} & \ci{0.48}{0.09} & \ci{0.50}{0.09}
 & \ci{0.53}{0.09} & \ci{0.59}{0.09} & \ci{0.53}{0.11}
 & \ci{0.39}{0.09} & \ci{\textbf{0.38}}{0.09} & \ci{0.58}{0.12}
 & 0 & \textbf{\$0.66} \\
\rowcolor{rowalt}No 3D
 & \ci{0.55}{0.05}
 & \ci{\textbf{0.65}}{0.07} & \ci{\textbf{0.65}}{0.08} & \ci{\textbf{0.76}}{0.07}
 & \ci{0.64}{0.07} & \ci{\textbf{0.65}}{0.08} & \ci{0.68}{0.09}
 & \ci{0.37}{0.09} & \ci{0.34}{0.08} & \ci{0.60}{0.14}
 & 60.8 & \$1.14 \\
No segmentation
 & \ci{0.55}{0.04}
 & \ci{0.59}{0.08} & \ci{0.58}{0.09} & \ci{0.68}{0.08}
 & \ci{\textbf{0.65}}{0.07} & \ci{\textbf{0.65}}{0.08} & \ci{\textbf{0.71}}{0.08}
 & \ci{0.42}{0.08} & \ci{0.37}{0.08} & \ci{\textbf{0.71}}{0.11}
 & 62.0 & \$1.36 \\
\rowcolor{rowbest}Full configuration
 & \ci{\textbf{0.56}}{0.05}
 & \ci{0.63}{0.08} & \ci{0.62}{0.09} & \ci{0.72}{0.08}
 & \ci{0.63}{0.08} & \ci{0.63}{0.09} & \ci{0.69}{0.10}
 & \ci{\textbf{0.43}}{0.08} & \ci{\textbf{0.38}}{0.08} & \ci{0.68}{0.13}
 & 59.6 & \$1.31 \\
\bottomrule
\end{tabular}
\end{adjustbox}
\endgroup
\end{table*}

\subsection{Main Results}
\label{sec:main_results}

\textbf{Off-the-shelf frontier VLMs are now capable of performing CHD classification through an interactive medical imaging environment.} With CHD-Harness, \gemini leads across all three cohorts, achieving macro F1 scores of 0.63 on ImageCHD and HVSMR-2.0 and 0.43 on 3DLabsCHD, with a mean of 0.56 (Table~\ref{tab:main}). \gptsol and \opus follow at 0.49, while the strongest open weight model, Kimi K3.1 Pro, reaches 0.31 with the same harness. The MedicalNet classifier reaches 0.26. This represents a substantial improvement over both the open weight VLMs and the trained volumetric classifier. More importantly, it demonstrates that a general-purpose VLM can perform complex anatomical diagnosis through interactive medical imaging software, without CHD-specific fine-tuning, complex prompts, or handcrafted diagnostic rules.

For reference, the rule-based baseline of Xu et al.~\cite{imagechd} achieves a macro recall of 0.72 on ImageCHD, after excluding AVSD, compared with 0.72 for \gemini. Differences in diagnostic annotations prevent a strictly equivalent comparison.

The weaker models reveal distinct limitations in their diagnostic behavior. Open weight VLMs generally exhibit lower recall than precision on the two public cohorts, indicating that they tend to miss existing defects. In contrast, the MedicalNet 3D classifier achieves higher recall than precision across all cohorts, frequently predicting abnormalities that are not present. Specialized medical VLMs perform particularly poorly: ClinFusion-8B predicts nearly every case as normal, reaching a mean macro F1 of 0.07, while M3D-LaMed achieves only 0.04. The strongest performance therefore comes from the general-purpose, closed weight VLMs that examine the scan interactively. Their advantage over the open weight VLMs using the same tools also shows how strongly performance depends on the underlying model's ability to interpret the anatomy.

The private 3DLabsCHD cohort presents the greatest challenge for the leading models, with all three closed weight VLMs experiencing substantial performance degradation. \gemini remains the strongest, achieving a macro F1 of 0.43, an 11-point advantage over the next-best model. Interestingly, its recall remains relatively high at 0.68, comparable to its performance on the public cohorts, while precision drops from 0.62 on ImageCHD to 0.38. This suggests that the primary challenge on this cohort is not simply detecting abnormalities, but accurately distinguishing which defects are present without overpredicting additional diagnoses. 

\begin{figure}[t]
  \centering
  \includegraphics[width=\columnwidth]{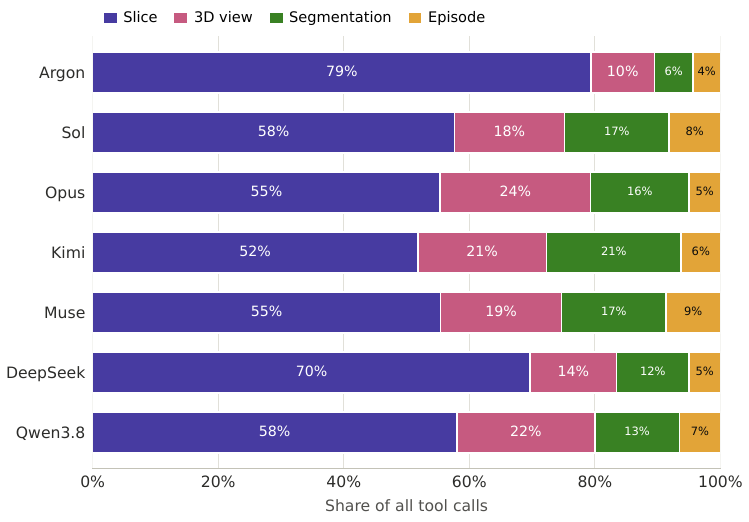}
  \caption{\textbf{Tool use across VLMs with CHD-Harness.} Share of tool calls over all 247 cases in each tool group by each VLM.}
  \label{fig:tool_calls}
\end{figure}

\begin{figure*}[t]
    \centering
    \includegraphics[width=\textwidth]{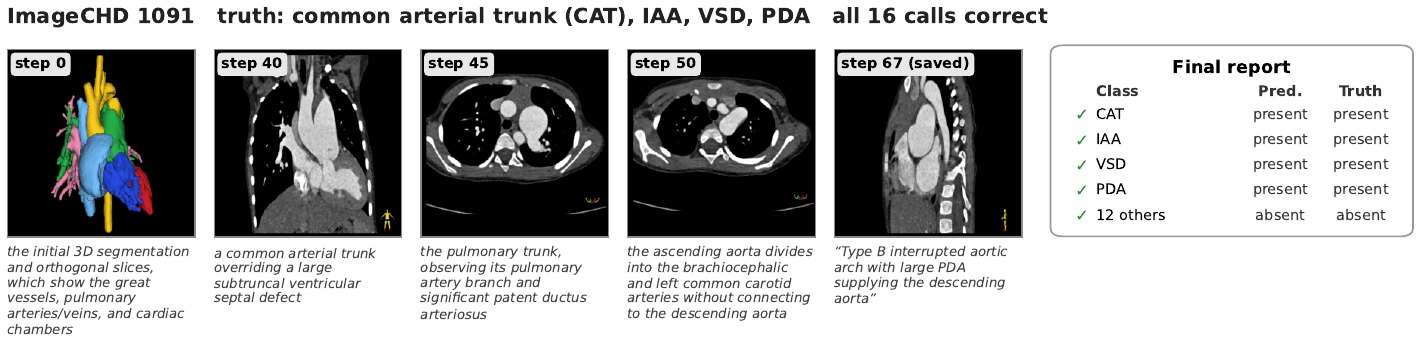}
    \caption{\textbf{Qualitative example of \gemini with CHD-Harness on an ImageCHD case.} Five views from the episode are shown with the final report against the ground truth. The text under each view is the model's own description of it, and step 67 is an example of bookmarked evidence.}
    \label{fig:qualitative}
\end{figure*}

The recorded examinations show how the VLMs use the imaging environment (Fig.~\ref{fig:tool_calls}). All seven models rely primarily on slice tools (52 to 79\% of calls), followed by 3D viewing and segmentation tools. \gemini and \gptsol make more tool calls on scans with defects than on normal scans, with median counts of 66 versus 39 and 61 versus 46, respectively. In contrast, \opus uses a median of 71 calls regardless of whether defects are present, approaching the 75-call limit, yet performs worse than \gemini while costing more than twice as much per case. Tool-call volume alone therefore does not explain the performance gap. In the ImageCHD example of Fig.~\ref{fig:qualitative}, \gemini traces the great arteries across the slices, identifies all four defects, and bookmarks the supporting view.

\subsection{Ablation Studies}
\label{sec:ablation}
Table~\ref{tab:ablation} compares the full configuration of \gemini with CHD-Harness against three ablations: removing segmentation, removing 3D tools, and replacing interactive examination with static views.

\noindent\textbf{Segmentation.}
To measure the effect of the segmentation, we remove the predicted masks of the six cardiac structures together with the five segmentation tools, leaving the model with the remaining 17 tools. Mean macro F1 decreases only marginally, from 0.56 to 0.55, indicating that CHD-Harness remains applicable even when no segmentation model is available. A notable exception is congenitally corrected transposition of the great arteries (ccTGA), in which the positions of the left and right ventricles are exchanged. Without segmentation, the number of correctly identified ccTGA cases on 3DLabsCHD decreases from 8/10 to 3/10, likely because the masks help the model identify each cardiac chamber.

\noindent\textbf{3D tools.}
To measure the effect of 3D viewing, we remove the 3D view together with the tools that operate on it. Mean macro F1 decreases only marginally, from 0.56 to 0.55. Recall increases by 4 points on ImageCHD and decreases by 1 and 8 points on HVSMR-2.0 and 3DLabsCHD, respectively, so the loss is concentrated in 3DLabsCHD. Within this cohort, the effect is largest for double-outlet right ventricle (DORV), in which both the aorta and the pulmonary artery arise predominantly from the right ventricle. The number of correctly identified DORV cases decreases from 26/43 to 18/43, likely because the origins of both great arteries are easier to trace in 3D than across slices.

\noindent\textbf{Interactive examination.}
To measure the effect of interactive examination, we remove all tools and present \gemini with each scan in a single request, using the same system prompt without its tool sections. The model can therefore neither select its views nor follow up on its observations. Instead, it receives axial, sagittal, and coronal slices spaced every 5\,mm with predicted segmentation outlines, along with static 3D views of the segmentation surfaces and volume rendering. Mean macro F1 decreases from 0.56 to 0.45, and recall decreases by 22, 16, and 10 points on ImageCHD, HVSMR-2.0, and 3DLabsCHD, respectively. The proportion of CHD cases classified as normal increases from 7\% to 16\%. Notably, the model with interactive tools achieves higher diagnostic performance while observing fewer images on average (81.6 versus 104.2 per case). 

Overall, interactive examination contributes most to diagnostic performance and improves detection across CHD types, whereas segmentation and 3D viewing benefit specific defects characterized by chamber identity and the course of the great arteries. 

\begin{figure}[t]
    \centering
    \includegraphics[width=\columnwidth]{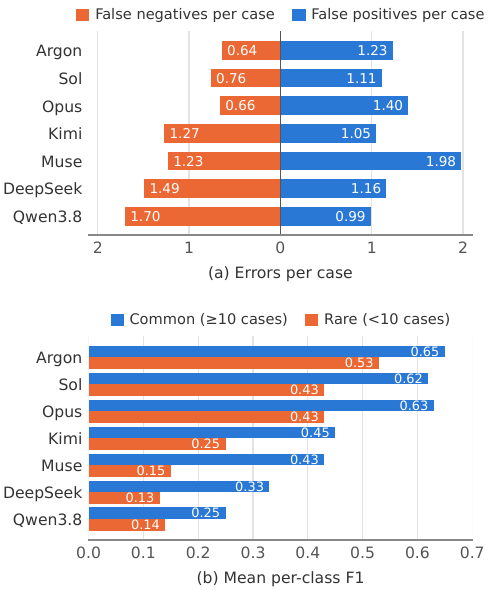}
    \caption{\textbf{Error patterns of the seven VLMs with CHD-Harness.} (a) Mean number of false negatives and false positives per case. (b) Mean per-class F1 on common defects ($\geq$10 positive cases) and rare defects ($<$10 positive cases).}
    \label{fig:error_patterns}
\end{figure}

\subsection{Error Analysis}
\label{sec:error}

\noindent\textbf{Error patterns across models.}
When related defects are grouped by the anatomical structures they affect (e.g., DORV and TGA as outflow tract defects), closed weight VLMs achieve a recall of 0.83--0.90, compared with 0.63--0.69 for individual defects. In contrast, open weight VLMs achieve a recall of only 0.46--0.68 on this grouping. They also fail to identify any true defect in 48--68\% of CHD cases, compared with 22--27\% for closed weight VLMs. Figure~\ref{fig:error_patterns}a shows the same difference in the errors per case. Open weight VLMs miss 1.23--1.70 defects per case, compared with 0.64--0.76 for closed weight VLMs, whereas six of the seven VLMs make a similar number of false positives (0.99--1.40 per case). These results suggest that closed weight VLMs more often confuse related defects, whereas open weight VLMs frequently miss abnormalities altogether.

Figure~\ref{fig:error_patterns}b shows that on common defects ($\geq$10 positive cases per cohort), the three closed weight VLMs achieve similar mean per-class F1 scores (0.62--0.65). However, Gemini 4 Argon performs better on rare defects ($<$10 positive cases), achieving 0.53 compared with 0.43 for GPT-6.1 Sol and Claude Opus 5.5.

\noindent\textbf{Failure mechanisms.}
Examining the traces demonstrates that models sometimes recognize possible defects but exclude them from their final diagnoses. For instance, Kimi K3.1 Pro mentioned 64\% of its false negatives during examination but omitted them from its final diagnosis. For DORV and TGA, models examined the ventricular origins of the great arteries but failed to assess how far they override the ventricular septum, an important distinction between these defects. Confidence is also unreliable: Argon's false-positive predictions have a median stated confidence of 0.95. These findings suggest that errors arise not only from false negatives but also from misinterpreting or dismissing observations.

\noindent\textbf{Confusion between related defects.}
Many false positives involve closely related heart defects. On 3DLabsCHD, 36 of \gemini's 44 false-positive VSD predictions occur in cases labeled as ToF or DORV. VSD is a component of ToF and is common in DORV~\cite{sts2025chsd,bellcheddar2023dorv}, so the dataset may record the broader diagnosis without listing VSD separately. On ImageCHD, 8 of 13 false-positive ToF predictions occur in cases with pulmonary atresia and VSD, a combination found in a severe form of ToF~\cite{tchervenkov2025pavsd}. Likewise, 17 of 22 false-positive TGA predictions on 3DLabsCHD occur in DORV cases. DORV can resemble TGA in the arrangement of the great arteries~\cite{bellcheddar2023dorv}, which helps explain why the model predicts TGA in these cases.

\section{Conclusion}
\label{sec:conclusion}
We introduced CHD-Bench, the first benchmark for interactive CHD diagnosis, defining the diagnostic task and evaluation protocol over 247 CT and MRI scans from three cohorts spanning 38 CHD types. We also introduced CHD-Harness, open-source software that runs a VLM in a 3D Slicer environment and records each examination. Among the models evaluated, \gemini with CHD-Harness achieved the highest performance across all three cohorts. Our experiments show that interactive examination improves diagnostic performance over non-interactive views, while segmentation masks provide a smaller overall benefit. These findings demonstrate the potential of VLMs for automated CHD diagnosis and highlight the promise of connecting general-purpose models to interactive medical imaging software. We release CHD-Bench and CHD-Harness to support reproducible evaluation and further research on interactive CHD diagnosis.

\noindent\textbf{Limitations.}
The clinical applicability of VLMs with CHD-Harness remains to be established through further validation with expert readers. Several CHD types are represented by only one to three cases, making their per-class results uncertain. Furthermore, exact-label evaluation penalizes clinically defensible diagnoses when diagnostic terminology differs from the dataset annotations. Finally, we evaluate each model only once per scan, leaving run-to-run variability unmeasured. 

\section*{Acknowledgments}
The authors would like to thank the Patrick J. McGovern Foundation and acknowledge its financial support for this research.

\section*{LLM Usage Statement}
We used OpenAI's ChatGPT and Anthropic's Claude as research and writing assistants throughout this work. Their use included support for research development, implementation, experimental analysis, visualizations, literature review, and manuscript preparation, including feedback on drafts. All AI-assisted contributions were reviewed and refined by the authors, who take full responsibility for the methodology, results, and final manuscript. No generative AI tools were used for dataset creation or annotation.

{
    \small
    \bibliographystyle{ieeenat_fullname}
    \bibliography{main}
}
\renewcommand{\topfraction}{0.9}
\renewcommand{\dbltopfraction}{0.9}
\renewcommand{\textfraction}{0.07}
\renewcommand{\floatpagefraction}{0.8}
\renewcommand{\dblfloatpagefraction}{0.8}
\setcounter{topnumber}{3}
\setcounter{dbltopnumber}{3}
\setcounter{totalnumber}{5}

\appendix
\twocolumn[{\centering\Large\textbf{Appendix}\par\vspace{1.0em}}]

\section{CHD-Bench Cohorts}
\label{sec:suppl_data}
\noindent\textbf{ImageCHD.} ImageCHD~\cite{imagechd} is a public dataset of 110 contrast-enhanced cardiac CT scans with case-level diagnoses over 16 CHD classes. We take the labels from the diagnosis sheet that the dataset authors updated on 10 May 2021, which supersedes the labels of the original publication. In the updated sheet, the AVSD column is computed as the logical AND of the ASD and VSD columns: every case with both defects is labeled AVSD. This rule does not reliably identify AVSD, a distinct abnormality involving the atrioventricular junction and valves. We therefore exclude AVSD from evaluation and score the remaining 15 classes. The scans were acquired on a Siemens Biograph 64 scanner.

\noindent\textbf{HVSMR-2.0.} HVSMR-2.0~\cite{hvsmr2} is a public dataset of 60 MRI scans with case-level clinical labels. These labels include CHD diagnoses alongside other information, such as prior surgery, other conditions, and imaging artifacts. We select the 20 existing labels that identify CHD types as our evaluation classes. We exclude labels for prior surgery (\eg, Glenn, Fontan), Marfan syndrome, imaging artifacts, and chamber dilation because they do not identify a CHD type. The dataset authors derived the labels from clinical records and verified them against each scan under the supervision of a pediatric cardiologist, excluding defects that had been corrected and were no longer visible. The scans were acquired at Boston Children's Hospital on a 1.50\,T Philips Achieva scanner, mostly with a free-breathing steady-state free precession sequence with respiratory navigation and prospective ECG gating.

\noindent\textbf{3DLabsCHD.} 3DLabsCHD is a private dataset of 86 cardiac CT scans collected at Amrita Institute of Medical Sciences, Kochi, India. Its diagnoses were assigned by a team of pediatric cardiologists led by a senior cardiologist with more than 20 years of experience, and they cover 15 CHD classes. It includes rare CHD types such as criss-cross heart, scimitar syndrome, and double-outlet left ventricle. The scans are contrast-enhanced and were acquired on scanners from several vendors (Philips, Siemens Healthineers, and GE Healthcare). All scans were de-identified before use.

\noindent\textbf{Exclusion criteria.} We use all scans where there are valid CHD labels. All 110 ImageCHD scans and all 86 3DLabsCHD scans are evaluated. In HVSMR-2.0, 9 of the 60 scans have no CHD labels, therefore they are not used. The diagnostic evaluation therefore includes 110 ImageCHD, 51 HVSMR-2.0, and 86 3DLabsCHD scans (247 total).

\noindent\textbf{Label harmonization.} The three cohorts contain 51 cohort-specific classes, including shared diagnoses and different names for the same defect. We harmonize these into a common set of labels. Double SVC (ImageCHD) and bilateral SVC (HVSMR-2.0) become double SVC; D-loop transposition (HVSMR-2.0) becomes TGA; and L-loop transposition (HVSMR-2.0) becomes congenitally corrected TGA, matching 3DLabsCHD. We also merge classes already named alike across cohorts, including ASD, VSD, DORV, ToF, DAA, and pulmonary atresia. This produces 39 distinct types, of which we evaluate 38 after excluding AVSD. Figure~\ref{fig:class_dist} shows their distribution. Common defects such as VSD and DORV appear in up to 62 cases in one cohort, while several types appear in only 1 to 3 cases.

\noindent\textbf{Label statistics.} The mean number of defects per abnormal case is 2.05 in ImageCHD, 1.37 in 3DLabsCHD and 4.58 in HVSMR-2.0 (43 abnormal cases), with up to 9 defects in one case. The full set of labels and their abbreviations are given in Table~\ref{tab:abbrev}.

\begin{table*}[t]
\centering
\caption{Abbreviations of the CHD classes in CHD-Bench.}
\label{tab:abbrev}
\footnotesize
\begin{tabular}{@{}ll@{\hspace{3em}}ll@{}}
\toprule
Abbreviation & CHD class & Abbreviation & CHD class \\
\midrule
AAH & Aortic arch hypoplasia & IAA & Interrupted aortic arch \\
AoAA & Aortic arch anomaly & InvA & Inverted atria \\
AoAn & Aortic aneurysm & InvV & Inverted ventricles \\
APVC & Anomalous pulmonary venous connection & LIVC & Left/central inferior vena cava \\
ASD & Atrial septal defect & LSVC & Left/central superior vena cava \\
ATS & Arterial tortuosity syndrome & MAPCAs & Major aortopulmonary collateral arteries \\
AVSD & Atrioventricular septal defect & MC & Mesocardia \\
CA & Common atrium & PA & Pulmonary atresia \\
CAT & Common arterial trunk & PA/MPAS & Pulmonary atresia or main pulmonary artery stump \\
ccTGA & Congenitally corrected transposition of the great arteries & PAS & Pulmonary artery sling \\
CCH & Criss-cross heart & PDA & Patent ductus arteriosus \\
CoA & Coarctation of the aorta & PS & Pulmonary stenosis \\
DAA & Double aortic arch & SIV & Superoinferior ventricles \\
DIDORV & Double-inlet double-outlet right ventricle & SS & Scimitar syndrome \\
DILV & Double-inlet left ventricle & SV & Single ventricle \\
DOLV & Double-outlet left ventricle & TGA & Transposition of the great arteries \\
DORV & Double-outlet right ventricle & ToF & Tetralogy of Fallot \\
DSVC & Double superior vena cava & TV & Tortuous vessels \\
DXC & Dextrocardia & VSD & Ventricular septal defect \\
HTX & Heterotaxy &  &  \\
\bottomrule
\end{tabular}
\end{table*}

\section{System Prompt}
\label{sec:suppl_prompt}
For each cohort, all interactive models use the same system prompt template. The ImageCHD template is reproduced in Listing~\ref{lst:system-prompt}. \texttt{\{max\_steps\}} is replaced by the tool-call budget (75). 

The HVSMR-2.0 prompt differs in three places: it names MRI instead of CT, it lists the 20 HVSMR-2.0 classes plus normal, and it adds a note that MRI intensities are signal values rather than Hounsfield units, so the HU window presets and HU readings do not have their CT meaning. In the setting without segmentation, the prompt omits the sentence on the loaded segmentation and the segmentation tools. The 3DLabsCHD prompt differs from the ImageCHD prompt only in its class list, which holds the 15 3DLabsCHD classes plus normal.

\begin{figure*}[p]
\begin{minipage}{\textwidth}
\begin{lstlisting}[
  style=chdprompt,
  caption={ImageCHD interactive system prompt template.},
  label={lst:system-prompt}
]
Congenital heart defect classification from cardiac CT

Role and task
You are an expert in cardiac imaging and congenital heart disease. You examine one
contrast-enhanced chest CT in a 3D viewer that you control with tools. Your task is to
classify the congenital heart defects in this patient. Use your own judgement: decide the best
path forward to meet your objective.

The patient can have no defect, one defect, or several. You have images only: no age, sex,
history or clinical information.

A segmentation of the heart is already loaded, with six structures: LV, RV, LA, RA, Ao and PA.
It is a guide, not ground truth, and it can be wrong.

The classes
Decide, for each of these 17 classes, whether it is present or not:

- atrial_septal_defect                           - double_aortic_arch
- ventricular_septal_defect                      - interrupted_aortic_arch
- atrioventricular_septal_defect                 - pulmonary_atresia
- tetralogy_of_fallot                            - anomalous_pulmonary_venous_connection
- transposition_of_great_arteries                - double_svc
- double_outlet_right_ventricle                  - patent_ductus_arteriosus
- common_arterial_trunk                          - pulmonary_artery_sling
- coarctation                                    - normal - no defects found.
- aortic_arch_hypoplasia

Mark every class that is present.

The tools
Each tool's description gives its arguments. In brief:

- Slices: set_slice, scroll_slice, set_window_preset, jump_to_point, point_at (reads the HU at
  a point), measure (distance in mm).
- 3D view: orbit, set_view_direction, zoom, pan_to, set_render_preset, set_clip_box,
  set_clip_plane, clear_clip.
- Segmentation display: set_segment_visibility, set_segment_opacity, set_3d_content,
  set_slice_outlines, center_on_segments.
- Episode: bookmark_view (saves the current views as evidence), reset_view, submit_report.

Coordinates are RAS (patient Right, Anterior, Superior) in millimetres. Each tool returns a
JSON block, then the images.

The episode
- You may make at most {max_steps} tool calls. calls_left in each tool result says how many
  are left.
- You may make several calls in one turn. They run in order.
- Before each call or batch, write one or two sentences: what you saw, and what you check
  next.
- Finish with one submit_report call. If it is rejected, read the reason, fix it, and submit
  again. Keep at least two calls for this.

Output
submit_report takes:
- classes: every one of the 17 classes, each with present (true or false) and confidence (0.5
  to 1, the probability that your call is correct). normal is present only when none of the
  other 16 is.
- evidence (optional): class name to a list of view refs from bookmark_view.
\end{lstlisting}
\end{minipage}
\end{figure*}

\section{Segmentation Model}
\label{sec:suppl_seg}
The six-structure masks that the model receives come from a SwinUNETR~\cite{swinunetr} (MONAI~\cite{monai} implementation, feature size 48, stochastic depth 0.15, one input channel). It has seven output channels, for the background, the left and right ventricles (LV, RV), the left and right atria (LA, RA), the aorta (Ao) and the pulmonary artery (PA). The network was trained from random initialization, and no pretrained weights were used.

We trained it by three-fold cross-validation on the three cohorts together. Each cohort was split into three folds on its own, with folds stratified by dataset and CHD type. The held-out cases per fold are 37/37/36 for ImageCHD, 20/20/20 for HVSMR-2.0 and 29/29/30 for 3DLabsCHD, which gives 258 cases in total. For each fold, one model was trained on the other two folds of all three cohorts, so every case's mask is predicted by a model that never trained on that case. Of the 258 scans used for segmentation, 247 also enter the diagnostic benchmark: 110 ImageCHD, 51 HVSMR-2.0, and 86 3DLabsCHD scans. The other 11 scans have segmentation labels but no valid CHD diagnosis. These are the 9 HVSMR-2.0 scans excluded in Sec.~\ref{sec:suppl_data} and two additional 3DLabsCHD scans that were annotated for segmentation only.

The training masks are the annotations released with ImageCHD and HVSMR-2.0, and the  annotations provided for 3DLabsCHD. ImageCHD and 3DLabsCHD share one label scheme (1 LV, 2 RV, 3 LA, 4 RA, 5 myocardium, 6 Ao, 7 PA). The HVSMR-2.0 labels (1 LV, 2 RV, 3 LA, 4 RA, 5 Ao, 6 PA, 7 SVC, 8 IVC, 9 myocardium) were mapped onto this scheme, with Ao moved from 5 to 6, PA from 6 to 7, and SVC and IVC set to background. Myocardium was not a training class.

Each scan was cropped to the heart region, reoriented to RAS, resized to $128 \times 128 \times 128$ voxels (trilinear interpolation for images, nearest-neighbor for labels), and z-score normalized over nonzero voxels. No data augmentation was applied. We trained the model using soft Dice loss (sigmoid outputs, squared denominator terms, background excluded), without cross-entropy or additional loss terms. Optimization used AdamW (learning rate $10^{-3}$, weight decay $10^{-4}$, betas 0.90 and $1-10^{-3}$), batch size 1, bfloat16 mixed precision, and gradient clipping at 1.00. The learning rate increased linearly from $10^{-4}$ over the first 10 epochs, followed by cosine decay to $10^{-6}$ over a maximum of 200 epochs. We validated every 2 epochs and stopped training if validation Dice did not improve for 40 epochs. For each fold, the model was trained on the other two folds, while the held-out fold was used for early stopping, checkpoint selection, and Dice reporting. The checkpoint with the highest validation Dice was used to generate the held-out masks, applying a sigmoid threshold of 0.50 and removing small disconnected components. The resulting masks were exported on the original full-scan grid for diagnosis. Training each fold took 3.10 to 4.50 hours on a single NVIDIA RTX 6000 Ada GPU. Across all 258 scans, the segmentation model achieved a mean Dice of $0.77 \pm 0.13$ (mean $\pm$ SD over cases), with per-cohort and per-structure results reported in Table~\ref{tab:seg_dice}.

\begin{table*}[t]
\centering
\caption{Held-out Dice of the segmentation model (mean $\pm$ SD over cases).}
\label{tab:seg_dice}
\footnotesize
\begin{tabular}{@{}lcccccccc@{}}
\toprule
Cohort & $n$ & LV & RV & LA & RA & Ao & PA & Mean \\
\midrule
ImageCHD & 110 & \ci{.84}{.15} & \ci{.81}{.15} & \ci{.84}{.13} & \ci{.85}{.12} & \ci{.84}{.14} & \ci{.71}{.17} & \ci{.82}{.11} \\
HVSMR-2.0 & 60 & \ci{.81}{.19} & \ci{.65}{.32} & \ci{.75}{.17} & \ci{.59}{.38} & \ci{.89}{.10} & \ci{.70}{.21} & \ci{.73}{.15} \\
3DLabsCHD & 88 & \ci{.77}{.21} & \ci{.73}{.19} & \ci{.76}{.15} & \ci{.73}{.21} & \ci{.78}{.16} & \ci{.62}{.19} & \ci{.73}{.12} \\
All & 258 & \ci{.81}{.19} & \ci{.74}{.22} & \ci{.79}{.15} & \ci{.75}{.25} & \ci{.83}{.14} & \ci{.68}{.19} & \ci{.77}{.13} \\
\bottomrule
\end{tabular}
\end{table*}

\section{Baselines}
\label{sec:suppl_baselines}

\noindent\textbf{Medical VLMs.} ClinFusion-8B~\cite{clinfusion}, M3D-LaMed~\cite{m3d} and NV-Reason-CT~\cite{nvreasonct} receive the full 3D volume once and produce one reply per case, with no tools. Their prompt follows the interactive template (Listing~\ref{lst:system-prompt}), with the sections about the viewer, the tools, the segmentation and the episode removed. The prompt keeps the role and task, the statement that a patient can have no defect, one or several, the dataset's class list, and the output format (one JSON object giving every class as present or absent, with a confidence). The prompt for each dataset differs only in the modality (CT or MRI) and the class list, and the ImageCHD prompt is reproduced in Listing~\ref{lst:volume-prompt}. These three baselines receive no segmentation masks.

\begin{figure}[tp]
\begin{minipage}{\linewidth}
\begin{lstlisting}[
  style=chdpromptcompact,
  caption={ImageCHD full-volume prompt for ClinFusion-8B, M3D-LaMed, and NV-Reason-CT.},
  label={lst:volume-prompt}
]
Congenital heart defect classification from
cardiac CT

Role and task
You are an expert in cardiac imaging and
congenital heart disease. You examine one
contrast-enhanced chest CT, given to you as
the full 3D volume. You get it once and have
no tools: you cannot change the view or ask
for anything more. Your task is to classify
the congenital heart defects in this patient.
Use your own judgement: decide the best path
forward to meet your objective.

The patient can have no defect, one defect, or
several. You have the CT only: no age, sex,
history or clinical information.

The classes
Decide, for each of these 17 classes, whether
it is present or not:

- atrial_septal_defect
- ventricular_septal_defect
- atrioventricular_septal_defect
- tetralogy_of_fallot
- transposition_of_great_arteries
- double_outlet_right_ventricle
- common_arterial_trunk
- coarctation
- aortic_arch_hypoplasia
- double_aortic_arch
- interrupted_aortic_arch
- pulmonary_atresia
- anomalous_pulmonary_venous_connection
- double_svc
- patent_ductus_arteriosus
- pulmonary_artery_sling
- normal - no defects found.

Mark every class that is present.

Output
End your reply with one JSON object:
- classes: every one of the 17 classes, each
  with present (true or false) and confidence
  (0.5 to 1, the probability that your call is
  correct). normal is present only when none
  of the other 16 is.

For example: {"classes":
{"atrial_septal_defect": {"present": false,
"confidence": 0.8}, ...}}
\end{lstlisting}
\end{minipage}
\end{figure}

\noindent\textbf{Response parsing for medical VLMs.} The three medical VLMs that receive full 3D volumes do not reliably follow the requested JSON format. To evaluate their diagnostic predictions despite these formatting failures, we implemented a rule-based parser that extracts predictions from both JSON and free-text replies. We apply the same parser to ClinFusion-8B, M3D-LaMed, and NV-Reason-CT. It first reads the JSON predictions, repairing minor brace errors. If no JSON object is found, it identifies defect names and checks for negation in the text. Classes without a recovered positive prediction count as absent; a reply with no recoverable prediction therefore counts as all classes absent. We assign Normal when no defect is predicted, rather than using the model's own Normal call.

\noindent\textbf{ClinFusion-8B.} We use the checkpoint ClinFusion-8B (Hugging Face commit \texttt{884c6ad8}, built on Qwen3-VL-8B-Instruct) with the model's own 3D input path. The volume is reoriented to RAS, resampled to 1\,mm, windowed to $[-1000, 1000]$\,HU and scaled to $[0, 1]$, then resized to $32 \times 256 \times 256$. Four 2D slices sampled from this volume are also given, as the model's adapter does by default. The prompt is the system turn and the user turn holds only the images. Decoding uses the checkpoint's own generation settings, which are sampling with temperature 0.70, top-k 20, top-p 0.80, seed 42 plus the case number, and at most 1,536 new tokens. This model has no reasoning options. It ran locally with Hugging Face Transformers in bfloat16 on one NVIDIA RTX 6000 Ada GPU. 

\noindent\textbf{M3D-LaMed.} We use the checkpoint M3D-LaMed-Phi-3-4B with the official CT preprocessing, which is raw voxel order, intensity clipping to the 0.05th--99.95th percentile of the foreground, z-scoring, min-max scaling to $[0, 1]$, foreground cropping and bilinear resizing to $32 \times 256 \times 256$. The input is the 256 image tokens followed by the prompt. M3D-LaMed has no chat template, so we end the prompt with ``\texttt{\textbackslash nAnswer:}''. Without this cue the model ends its reply at once. Decoding is greedy, with at most 1,536 new tokens. This model has no reasoning options. It ran locally with Hugging Face Transformers in bfloat16 on one NVIDIA RTX 6000 Ada GPU.

\noindent\textbf{NV-Reason-CT.} We use the checkpoint \texttt{nvidia/NV-Reason-CT} (revision \texttt{386b93e0}). The volume is given in HU with a correct header and goes through the model's own processor, which applies LPS orientation, 2\,mm isotropic resampling, padding with $-1000$\,HU to $192^3$, a centred $192^3$ crop (cardiac fields of view are smaller than 384\,mm, so the whole scan fits) and clipping to $[-1000, 1000]$\,HU. The message is one user turn holding the volume and the prompt. Decoding is greedy with the model's thinking mode on (its default), with at most 8,192 new tokens. It ran locally with Hugging Face Transformers 5.6.2 in bfloat16 on one NVIDIA RTX 6000 Ada GPU. 

The NV-Reason-CT and ClinFusion-8B pipelines expect CT intensities in the range $[-1000, 1000]$\,HU. The HVSMR-2.0 MRI scans use a different intensity scale. For these two models, we therefore linearly map each MRI volume's 0.50th--99.50th intensity percentile range to $[-1000, 1000]$ before passing it to the pipeline.

\noindent\textbf{MedGemma-27B.} We use the checkpoint \texttt{google/medgemma-27b-it} (revision \texttt{2d3e00ea})~\cite{medgemma}. MedGemma-27B is a 2D model, so it received exactly the images of the static-view ablation (Sec.~\ref{sec:suppl_static}), in the same order and with the same system prompt (Listing~\ref{lst:static-prompt}). These are axial, sagittal and coronal slices every 5\,mm with the segmentation outlined, plus the volume rendering and the segmentation shells seen from six sides (on average 104.20 images per case, with 107.10 for ImageCHD, 101.20 for HVSMR-2.0 and 102.20 for 3DLabsCHD). Its reply is parsed using the same report format as the interactive VLMs (one JSON object with every class). MedGemma-27B does not produce a valid report for 56 of the 247 cases (29 in ImageCHD, 4 in HVSMR-2.0 and 23 in 3DLabsCHD), and these cases are excluded from its metrics. Decoding uses temperature 0 and at most 32,000 output tokens, and MedGemma has no reasoning mode. It ran locally on two NVIDIA RTX 6000 Ada GPUs with a 131,072-token context.

\noindent\textbf{Volumetric classifier.} Each scan is resized to $128^3$ voxels, and the features of the last stage of the frozen MedicalNet ResNet-18~\cite{chen2019med3d} are max-pooled into a 512-dimensional vector. Class-balanced linear SVMs with $C = 10^{-2}$ then predict the labels, one SVM per label, trained separately for each cohort on that cohort's own labels. We use the same three folds as the segmentation model within each cohort. We standardize the features with the mean and standard deviation of the training folds only.

\noindent\textbf{Interactive VLMs.} Table~\ref{tab:suppl_interactive_settings} lists the model IDs, API routes, reasoning effort requested by the runner, and per-turn output caps for the interactive models. We do not set the sampling temperature, so the default of each provider applies. Images are sent losslessly. \opus is accessed through the Anthropic API and all other interactive VLMs through OpenRouter.

\section{Evaluation and Code}
\label{sec:suppl_code}
The code is publicly available at \url{https://github.com/Maitri-Labs-Amrita/CHD-Bench}. Its README provides the instructions for running the code.

We compute precision, recall, and F1 separately for each class and report their macro averages within each cohort, giving every class equal weight. When a model predicts no positive cases for a class, precision is undefined. We set it to zero so that failing to predict a class does not remove that class from the average.

The ImageCHD prompt includes AVSD, but we exclude this class from the macro averages because its reference labels do not reliably identify AVSD (Sec.~\ref{sec:suppl_data}). We still count an AVSD prediction as a predicted defect when determining whether a case is Normal: any predicted defect, including AVSD, rules out Normal.

To quantify uncertainty due to the finite number of patients, we draw 2{,}000 bootstrap resamples within each cohort by sampling patients with replacement, keeping each patient's labels and predictions together. A rare class can have no ground-truth positives in a resample. We omit such classes from that resample's macro averages because recall cannot be evaluated without positive examples. The main tables report 95\% confidence intervals estimated from these resamples. The released batch scorer uses a normal approximation: it estimates the standard error from the variation in scores across resamples and reports the original cohort score $\pm$ 1.96 times this standard error.

\begin{table*}[t]
\centering
\caption{Interactive model settings. Reasoning effort is the setting requested by the runner, and each model uses its highest available level (\texttt{high} is the highest level for Gemini 4 Argon). The standard output cap was 32,000 tokens per turn. Two retries each for DeepSeek and Qwen used 64,000 tokens per turn.}
\label{tab:suppl_interactive_settings}
\small
\begin{tabular}{@{}lllc r@{}}
\toprule
Model & Model ID & API route & Reasoning effort & Output cap (tokens) \\
\midrule
Gemini 4 Argon & \texttt{google/gemini-4-argon} & OpenRouter & \texttt{high} & 32,000 \\
GPT-6.1 Sol & \texttt{openai/gpt-6.1-sol} & OpenRouter & \texttt{max} & 32,000 \\
Claude Opus 5.5 & \texttt{claude-opus-5-5} & Anthropic & \texttt{max} & 32,000 \\
Kimi K3.1 Pro & \texttt{moonshotai/kimi-k3.1-pro} & OpenRouter & \texttt{max} & 32,000 \\
Muse Spark Open & \texttt{meta/muse-spark-open} & OpenRouter & \texttt{max} & 32,000 \\
DeepSeek V4.1 Flash & \texttt{deepseek/deepseek-v4.1-flash} & OpenRouter & \texttt{max} & 32,000 / 64,000 \\
Qwen3.8-27B & \texttt{qwen/qwen3.8-27b} & OpenRouter & \texttt{max} & 32,000 / 64,000 \\
\bottomrule
\end{tabular}
\end{table*}

\section{Static-View Setting}
\label{sec:suppl_static}
\begin{figure*}[t]
    \centering
    \includegraphics[width=\textwidth]{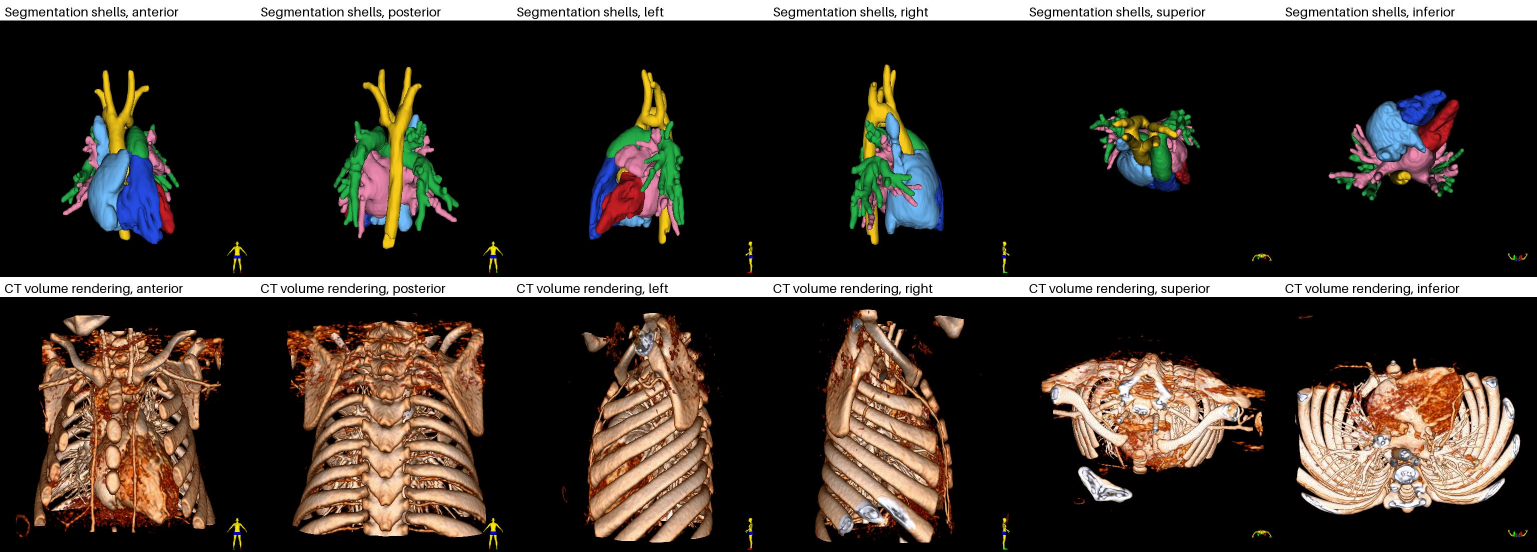}
    \caption{\textbf{The twelve 3D views of the static-view setting for one ImageCHD scan (case 1036).} Top: the segmentation shells (LA pink, RA light blue, LV dark red, RV dark blue, Ao yellow, PA green). Bottom: the volume rendering with the \textit{CT-Coronary-Arteries-2} preset. Each row shows the anterior, posterior, left, right, superior and inferior views.}
    \label{fig:static_3d}
\end{figure*}

In the static-view ablation, the model receives a fixed image set in a single turn and has no tools. Every image is rendered by the 3D Slicer environment used by CHD-Harness at $512 \times 512$ pixels and sent without cropping, downscaling or lossy compression. The initial output limit was 32{,}000 tokens. After reaching that limit, one ImageCHD case and one 3DLabsCHD case were resent with a 64{,}000-token limit.

\noindent\textbf{Slices.} The slices cover the bounding box of the predicted segmentation, extended by 40\,mm superiorly and inferiorly and by 20\,mm in the other four directions and clipped to the scan. Axial, sagittal and coronal slices are placed every 5\,mm, centered in this range, so the spacing is the same for every scan and only the number of slices varies. Table~\ref{tab:static_counts} reports the minimum, median, and maximum number of images per scan for each viewing plane across the three cohorts. Each image shows the whole slice plane, as in the interactive setting, in the environment's default grayscale window (width 350\,HU and level 40\,HU for CT, width 1000 and level 500 for MRI). Each of the six predicted structures is outlined in its color where the slice cuts it, without fill.

\noindent\textbf{3D views.} Six images show the segmentation alone, using the same colors and initial 3D view as the interactive setting. Each structure is drawn as a shell 1\,mm thick around its segmented blood, and the shell is opaque, so these images show the outer surface of each structure. Six more show the volume rendering alone, with the \textit{CT-Coronary-Arteries-2} preset of 3D Slicer for CT and the\textit{ MR-BloodPool} preset for MRI. Both sets are taken from the anterior, posterior, left, right, superior and inferior sides with a perspective camera that looks at the center of the volume from a fixed distance at which the whole volume is in view. Figure~\ref{fig:static_3d} shows the twelve 3D views of one ImageCHD scan.

\noindent\textbf{Prompt.} Listing~\ref{lst:static-prompt} shows the ImageCHD system prompt used by the static-view ablation and MedGemma-27B. It is the interactive prompt (Listing~\ref{lst:system-prompt}) with the tool sections replaced by a description of the images. For HVSMR-2.0 and 3DLabsCHD, only the modality words and the class list change. In the listing, \texttt{\{current\_cohort\_classes\}} stands for the class list inserted for the current cohort. The user message starts with the geometry block that the model receives at its first step (volume size, voxel spacing, extent in patient coordinates, and the names and colors of the segments). The message before each slice gives its plane, its index in the stack and its absolute position in millimeters along the slice axis, which is the coordinate that the model passes to \texttt{set\_slice}. The message before each 3D image contains its content and its view direction. The model answers with one present/absent decision per class, as in the interactive setting.

\begin{figure}[tp]
\begin{minipage}{\linewidth}
\begin{lstlisting}[
  style=chdpromptcompact,
  caption={ImageCHD static-view prompt for MedGemma-27B and the static-view ablation.},
  label={lst:static-prompt}
]
Congenital heart defect classification from
cardiac CT

Role and task
You are an expert in cardiac imaging and
congenital heart disease. You examine one
contrast-enhanced chest CT from a fixed set of
images taken in a 3D viewer. You get these
images once and have no tools: you cannot
change the views or ask for more images. Your
task is to classify the congenital heart
defects in this patient. Use your own
judgement: decide the best path forward to
meet your objective.

The patient can have no defect, one defect, or
several. You have images only: no age, sex,
history or clinical information.

A segmentation of the heart is already loaded,
with six structures: LV, RV, LA, RA, Ao and
PA. It is a guide, not ground truth, and it
can be wrong.

The classes
{current_cohort_classes}

Mark every class that is present.

The images
- Slices: axial, sagittal and coronal slices,
  evenly spaced every 5 mm, window
  CT-Mediastinum. Each structure's segmented
  blood is outlined in its colour where the
  slice cuts it (LA pink, RA light blue, LV
  dark red, RV dark blue, Ao yellow, PA
  green).
- 3D view: the CT volume rendering (preset
  CT-Coronary-Arteries-2) and the structure
  shells alone, each seen from six sides:
  anterior (superior up), posterior (superior
  up), left (superior up), right (superior
  up), superior (posterior up), inferior
  (anterior up).

Coordinates are RAS (patient Right, Anterior,
Superior) in millimetres. A slice's position
is along its own axis: axial -> S (larger =
more superior), sagittal -> R (larger =
further to the patient's right), coronal -> A
(larger = more anterior). A line of text
before each image says which image it is.

Output
End your reply with one JSON object:
- classes: every one of the 17 classes, each
  with present (true or false) and confidence
  (0.5 to 1, the probability that your call is
  correct). normal is present only when none
  of the other 16 is.

For example: {"classes":
{"atrial_septal_defect": {"present": false,
"confidence": 0.8}, ...}}
\end{lstlisting}
\end{minipage}
\end{figure}

The user turn starts with ``The case is loaded. These are all the images you get.'', then gives the scan geometry, and then each image after a one-line label such as ``[axial 1 of 34, S = 165.0 mm]''.

\begin{table}[t]
\centering
\caption{Images per scan in the static-view setting, as minimum / median / maximum. The total includes the twelve 3D views.}
\label{tab:static_counts}
\footnotesize
\setlength{\tabcolsep}{2pt}
\begin{tabular}{@{}lcccc@{}}
\toprule
Cohort & Axial & Sagittal & Coronal & Total \\
\midrule
ImageCHD & 21\,/\,42\,/\,106 & 15\,/\,29\,/\,53 & 14\,/\,24\,/\,39 & 68\,/\,105.50\,/\,199 \\
HVSMR-2.0 & 24\,/\,36\,/\,53 & 13\,/\,25\,/\,40 & 18\,/\,24\,/\,39 & 73\,/\,97\,/\,134 \\
3DLabsCHD & 21\,/\,35\,/\,64 & 16\,/\,28.50\,/\,44 & 18\,/\,24.50\,/\,38 & 74\,/\,101.50\,/\,150 \\
\bottomrule
\end{tabular}
\end{table}

\section{Ablation Without 3D Tools}
\label{sec:suppl_no3d}
The ``No 3D'' ablation keeps the main setup, including the preloaded segmentation, but removes everything three-dimensional. Ten of the 22 tools are removed. These are the 3D-view tools \texttt{orbit}, \texttt{set\_\allowbreak{}view\_\allowbreak{}direction}, \texttt{pan\_\allowbreak{}to}, \texttt{set\_\allowbreak{}render\_\allowbreak{}preset}, \texttt{set\_\allowbreak{}clip\_\allowbreak{}box}, \texttt{set\_\allowbreak{}clip\_\allowbreak{}plane} and \texttt{clear\_\allowbreak{}clip}, and the 3D segmentation-display tools \texttt{set\_\allowbreak{}segment\_\allowbreak{}opacity}, \texttt{set\_\allowbreak{}3d\_\allowbreak{}content} and \texttt{center\_\allowbreak{}on\_\allowbreak{}segments}. The model keeps 12 tools, which are \texttt{set\_\allowbreak{}slice}, \texttt{scroll\_\allowbreak{}slice}, \texttt{set\_\allowbreak{}window\_\allowbreak{}preset}, \texttt{jump\_\allowbreak{}to\_\allowbreak{}point}, \texttt{point\_\allowbreak{}at}, \texttt{zoom}, \texttt{measure}, \texttt{set\_\allowbreak{}segment\_\allowbreak{}visibility}, \texttt{set\_\allowbreak{}slice\_\allowbreak{}outlines}, \texttt{bookmark\_\allowbreak{}view}, \texttt{reset\_\allowbreak{}view} and \texttt{submit\_\allowbreak{}report}. Every observation, including the initial one, holds only the axial, sagittal and coronal slice views. The model never receives a 3D image. 

\noindent\textbf{Prompt changes.} The interactive system prompt (Listing~\ref{lst:system-prompt}) changes in two places. ``a 3D viewer that you control with tools'' becomes ``a viewer that you control with tools''. In the tool summary, the line listing the 3D-view tools is removed, \texttt{zoom} moves to the slice tools, and the segmentation-display line lists only \texttt{set\_\allowbreak{}segment\_\allowbreak{}visibility} and \texttt{set\_\allowbreak{}slice\_\allowbreak{}outlines}. All other settings match the main \gemini runs (reasoning effort high, 75 tool calls, USD 10 cost cap per episode, 32,000 output tokens per turn).

\section{Per-Class Results}
\label{sec:suppl_perclass}
\begin{figure*}[t]
    \centering
    \includegraphics[width=\textwidth]{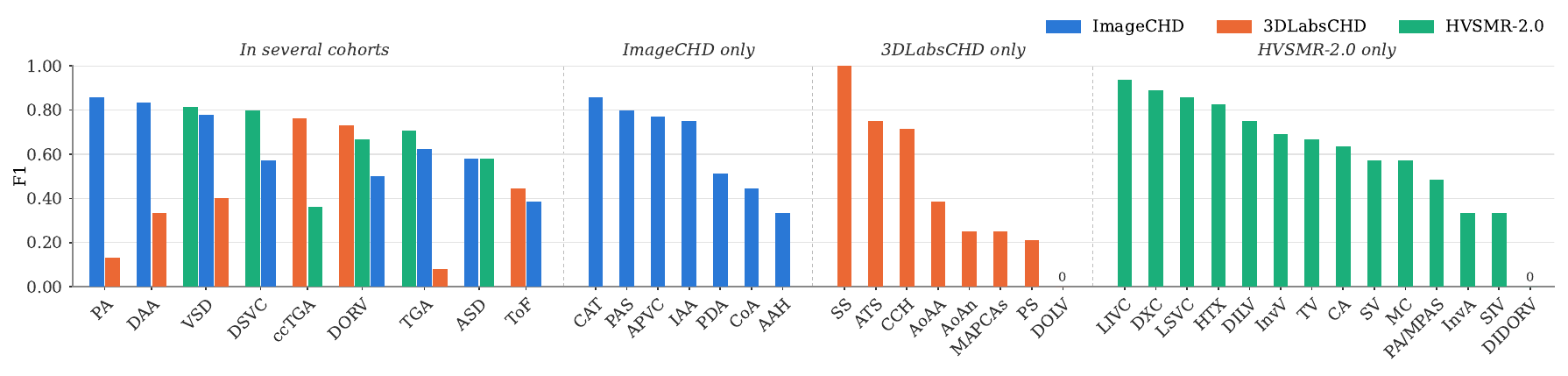}
    \caption{\textbf{Per-class F1 scores of \gemini across the three cohorts.} Classes are grouped by cohort coverage and ordered by mean F1 for shared classes and by F1 for cohort-specific classes. Missing bars indicate classes not evaluated in a cohort. The class abbreviations are defined in Table~\ref{tab:abbrev}.}
    \label{fig:perclass}
\end{figure*}

\begin{table*}[t]
\centering
\caption{Recall by anatomical group across all three cohorts. Each positive case-label pair contributes once per model.}
\label{tab:suppl_anatomical_recall}
\small
\begin{tabular}{@{}lccccccc@{}}
\toprule
Group & Gemini & Sol & Opus & Kimi & DeepSeek & Qwen & Muse \\
\midrule
Arch and arteries (52) & \textbf{\ci{.75}{.11}} & \ci{.52}{.15} & \ci{.58}{.14} & \ci{.35}{.14} & \ci{.17}{.11} & \ci{.23}{.13} & \ci{.19}{.11} \\
Systemic and pulmonary veins (49) & \textbf{\ci{.86}{.11}} & \ci{.67}{.13} & \ci{.76}{.14} & \ci{.22}{.15} & \ci{.12}{.09} & \ci{.12}{.11} & \ci{.37}{.16} \\
Outflow (164) & \textbf{\ci{.64}{.08}} & \ci{.62}{.08} & \ci{.59}{.08} & \ci{.46}{.08} & \ci{.24}{.06} & \ci{.21}{.06} & \ci{.46}{.08} \\
Septal (183) & \ci{.66}{.07} & \ci{.67}{.07} & \textbf{\ci{.79}{.05}} & \ci{.37}{.06} & \ci{.41}{.07} & \ci{.10}{.04} & \ci{.43}{.07} \\
Position, situs, and ventricular configuration (66) & \textbf{\ci{.73}{.13}} & \ci{.62}{.14} & \ci{.65}{.13} & \ci{.42}{.13} & \ci{.27}{.11} & \ci{.33}{.11} & \ci{.44}{.15} \\
\bottomrule
\end{tabular}
\end{table*}

\begin{table*}[t]
\centering
\caption{CHD classes included in each anatomical group. Class abbreviations are defined in Table~\ref{tab:abbrev}.}
\label{tab:suppl_anatomical_groups}
\small
\begin{tabular}{@{}p{0.34\textwidth}p{0.60\textwidth}@{}}
\toprule
Anatomical group & CHD classes \\
\midrule
Septal defects & ASD, VSD, CA \\
Outflow defects & ToF, TGA, ccTGA, DORV, CAT, PA, DOLV, PA/MPAS, PS, DIDORV \\
Arch and arterial defects & CoA, AAH, DAA, IAA, PDA, PAS, AoAA, AoAn, ATS, MAPCAs, TV \\
Systemic and pulmonary venous anomalies & APVC, DSVC, LSVC, LIVC, SS \\
Position, situs, and ventricular configuration & DXC, MC, HTX, InvA, InvV, SIV, CCH, SV, DILV \\
\bottomrule
\end{tabular}
\end{table*}

Figure~\ref{fig:perclass} shows the F1 score of \gemini with CHD-Harness for every class of the three cohorts. Gemini scored highest on classes defined by an abnormal vessel course or heart position, such as left or central IVC (0.94) and dextrocardia (0.89) in HVSMR-2.0, common arterial trunk (0.86) in ImageCHD and scimitar syndrome (1.00) in 3DLabsCHD. Most low scores came from false positives rather than missed cases, for example for ToF and coarctation in ImageCHD and for TGA, pulmonary atresia and pulmonary stenosis in 3DLabsCHD, where TGA and pulmonary atresia scored 0.08 and 0.13 although the same classes scored 0.62 and 0.86 in ImageCHD. ASD showed the opposite pattern, with about half of its cases missed in ImageCHD and HVSMR-2.0.

\section{Exploratory Analysis of Examination Behavior}
\label{sec:suppl_behavior}
We analyzed the recorded examinations of the seven interactive VLMs across the three cohorts (1,729 runs). These post hoc analyses describe associations in the recorded runs.

\subsection{Recall by anatomical group}
Defects that involve the same part of the heart, such as the aortic arch or the venous connections, require the model to examine similar anatomy. We therefore grouped related defects by anatomy and computed each model's recall within each group, to see which kinds of defect each model identifies well.
Table~\ref{tab:suppl_anatomical_recall} reports recall across five anatomical groups, pooling positive case-label pairs from all three cohorts. A prediction is counted as correct only when it matches the specific CHD class, rather than another class in the same group. Table~\ref{tab:suppl_anatomical_groups} lists the classes included in each group. Gemini achieved the highest recall for arch and arterial defects (0.75) and venous anomalies (0.86), while Opus performed best on septal defects (0.79).

\begin{table}
\centering
\caption{Median distinct slices through the heart and mean per-case F1 across the three cohorts.}
\label{tab:suppl_coverage_behavior}
\small
\begin{tabular}{@{}lcc@{}}
\toprule
Model & Heart slices & Per-case F1 \\
\midrule
Gemini 4 Argon & 40 & .57 \\
GPT-6.1 Sol & 37 & .53 \\
Claude Opus 5.5 & 42 & .53 \\
Kimi K3.1 Pro & 27 & .37 \\
DeepSeek V4.1 Flash & 50 & .28 \\
Qwen3.8-27B & 23 & .21 \\
Muse Spark Open & 22 & .28 \\
\bottomrule
\end{tabular}
\end{table}

\begin{table*}[t]
\centering
\caption{Trivial baselines scored with the metrics of Sec.~\ref{sec:chd_bench}. Prevalence-random gives the mean over 1,000 draws.}
\label{tab:suppl_trivial}
\footnotesize
\begin{tabular}{@{}lcccccccccc@{}}
\toprule
 & & \multicolumn{3}{c}{ImageCHD} & \multicolumn{3}{c}{HVSMR-2.0} & \multicolumn{3}{c}{3DLabsCHD} \\
\cmidrule(lr){3-5}\cmidrule(lr){6-8}\cmidrule(l){9-11}
Baseline & Mean F1 & F1 & P & R & F1 & P & R & F1 & P & R \\
\midrule
All-present & 0.20 & 0.18 & 0.11 & 0.94 & 0.29 & 0.18 & 0.95 & 0.15 & 0.09 & 1.00 \\
Prevalence-random & 0.13 & 0.12 & 0.12 & 0.12 & 0.18 & 0.19 & 0.18 & 0.09 & 0.09 & 0.09 \\
\bottomrule
\end{tabular}
\end{table*}

\begin{table*}[t]
\centering
\caption{Paired differences in macro F1 (first minus second) with 95\% intervals and two-sided $p$-values from 2,000 paired patient resamples.}
\label{tab:suppl_paired}
\footnotesize
\begin{tabular}{@{}lcccccccc@{}}
\toprule
 & \multicolumn{2}{c}{ImageCHD} & \multicolumn{2}{c}{HVSMR-2.0} & \multicolumn{2}{c}{3DLabsCHD} & \multicolumn{2}{c}{Mean} \\
\cmidrule(lr){2-3}\cmidrule(lr){4-5}\cmidrule(lr){6-7}\cmidrule(l){8-9}
Comparison & $\Delta$F1 & $p$ & $\Delta$F1 & $p$ & $\Delta$F1 & $p$ & $\Delta$F1 & $p$ \\
\midrule
\gemini vs.\ \gptsol & \ci{0.06}{0.09} & 0.185 & \ci{0.06}{0.07} & 0.181 & \ci{0.12}{0.09} & 0.006 & \ci{0.08}{0.05} & 0.001 \\
\gemini vs.\ \opus & \ci{0.06}{0.07} & 0.065 & \ci{0.06}{0.07} & 0.111 & \ci{0.11}{0.09} & 0.011 & \ci{0.08}{0.05} & $<$0.001 \\
\midrule
\textit{\gemini} & & & & & & & &\\
Full setup vs.\ Static views & \ci{0.20}{0.09} & $<$0.001 & \ci{0.10}{0.08} & 0.016 & \ci{0.04}{0.08} & 0.220 & \ci{0.11}{0.05} & $<$0.001 \\
Full setup vs.\ No 3D & \ci{$-$0.02}{0.07} & 0.521 & \ci{$-$0.01}{0.08} & 0.827 & \ci{0.06}{0.05} & $<$0.001 & \ci{0.01}{0.04} & 0.529 \\
Full setup vs.\ No segmentation & \ci{0.04}{0.05} & 0.202 & \ci{$-$0.02}{0.07} & 0.524 & \ci{0.01}{0.06} & 0.766 & \ci{0.01}{0.04} & 0.675 \\
\bottomrule
\end{tabular}
\end{table*}

\subsection{Slice coverage and diagnostic performance}
We examined whether models that inspect more of the heart achieve better diagnostic performance. Using the recorded tool calls, we counted the distinct axial, sagittal, and coronal slices viewed within anatomical regions defined by segmentation bounding boxes. This provides an approximate measure of how extensively each model examined the anatomy. Table~\ref{tab:suppl_coverage_behavior} compares the median number of distinct heart slices viewed by each model with its mean per-case F1, calculated for each scan and then averaged across scans. Notably, DeepSeek examined the most slices (median 50) but achieved a mean per-case F1 of only 0.28, compared with 0.57 for Gemini, which examined a median of 40 slices.

We further tested whether examining more slices in a particular anatomical region was associated with correctly identifying defects in that region. We used the five anatomical groups defined above to measure coverage, but evaluated detection using the individual CHD labels. After accounting for differences between models and cases, we found no statistically significant association between regional slice coverage and correct defect detection ($p=0.14$ to $0.68$ across the five groups). For defects involving the aortic arch and veins, Gemini examined fewer relevant slices than Sol and Opus (median 28 versus 35 for both), yet achieved higher recall. After accounting for differences between CHD classes, Gemini's estimated recall advantage over Sol and Opus was 17.3 percentage points (95\% CI 7.3--27.4). This advantage remained at 15.1 percentage points after additionally accounting for regional slice coverage (95\% CI 4.3--25.9; 303 model-label observations).

Finally, we examined whether viewing more slices was associated with fewer diagnostic errors. Across runs, greater slice coverage was associated with more false-positive predictions (0.42 additional false-positive labels per unit increase in $\log(1+\text{distinct slices})$, $p<10^{-3}$), but not with a significant reduction in missed defects (0.03 fewer false negatives, $p=0.72$). These comparisons account for differences between models, cohorts, and the number of actual defects in each scan. Together, these findings suggest that examining more slices does not necessarily improve diagnostic accuracy. In particular, Gemini's advantage over Sol and Opus persists even after accounting for how extensively they examined the relevant anatomy, suggesting that diagnostic performance depends on more than the amount of anatomy inspected.

\subsection{Model-specific examination patterns}
Gemini averaged 5.50 tool calls per turn. It made approximately 3.50 additional calls per extra reference defect. Sol usually made one tool call per turn. Opus used a median of 71 of the 75 available calls and reported at least one defect on 52\% of cases scored as Normal. It had the highest family-level recall (0.90), which gives credit when the model predicts any defect from the correct anatomical group. Kimi discussed 64\% of its missed labels in its visible reasoning before the final report. This text-based estimate includes negated mentions. DeepSeek reached the 75-call budget in 33 of 247 runs, and intensity probes accounted for 16\% of its tool calls. Muse averaged 1.98 false-positive labels per case and reported a defect on 67\% of cases scored as Normal. In Muse's visible reasoning, 73\% of its reported classes were first discussed during the last fifth of the run. Qwen used a median of 39 tool calls and reported an average of 1.37 defects per case, compared with 2.08 reference defects.

\section{Trivial Baselines}
\label{sec:suppl_trivial}
To put the reported macro F1 scores in context, Table~\ref{tab:suppl_trivial} evaluates two baselines that make predictions without examining the scan, using the released batch scorer. The all-present baseline calls every defect class present and therefore never predicts Normal. The prevalence-random baseline calls each defect class present with probability equal to its prevalence in the cohort and predicts Normal when it calls no defect, and the table reports its mean over 1,000 random draws. 
A baseline that ignores the scan already reaches a mean macro F1 of 0.20, because frequent classes such as VSD and DORV skew the results favorably for these baselines. Macro F1 therefore reflects diagnostic information only to the extent that it exceeds these baselines. The three closed-weight VLMs and Kimi K3.1 Pro scored above both baselines in every cohort, whereas Muse Spark Open, DeepSeek V4.1 Flash and Qwen3.8-27B reached mean F1 values between 0.18 and 0.23, close to the all-present baseline, and the four medical VLMs scored below the prevalence-random baseline in mean F1.

\section{Paired Comparisons}
\label{sec:suppl_paired}
Several differences in macro F1 in Secs.~\ref{sec:main_results} and~\ref{sec:ablation} are small, and confidence intervals for individual systems do not directly quantify uncertainty in these differences. We therefore compare systems using a paired patient-level bootstrap (Table~\ref{tab:suppl_paired}). Each of 2,000 resamples draws the patients of each cohort with replacement and scores both systems on the same patients. The table gives the difference in macro F1 with its 95\% interval (point $\pm$ 1.96 times the bootstrap standard error) and a two-sided $p$-value, which is twice the smaller share of resampled differences on either side of zero. The mean over cohorts averages the three cohort differences in each resample. \gemini scored higher than \gptsol and \opus on the mean over cohorts, and its lead was largest on 3DLabsCHD, while the differences within ImageCHD or HVSMR-2.0 alone stayed within the bootstrap uncertainty. Interactive examination scored higher than static views on the mean, with the largest gain on ImageCHD and no clear gain on 3DLabsCHD. Removing the 3D view or the segmentation changed the mean macro F1 by about 0.01, in line with Sec.~\ref{sec:ablation}, although removing the 3D view lowered F1 on 3DLabsCHD by 0.06.

\end{document}